\documentclass{article}

\usepackage[final]{neurips_2024}
\usepackage{amsmath}
\usepackage{amssymb}

\usepackage[utf8]{inputenc} 
\usepackage[T1]{fontenc}    
\usepackage{hyperref}       
\usepackage{url}            
\usepackage{booktabs}       
\usepackage{amsfonts}       
\usepackage{nicefrac}       
\usepackage{microtype}      
\usepackage{xcolor}         
\usepackage{graphicx} 
\usepackage{tabularx}
\usepackage{lipsum}
\usepackage{multirow}
\usepackage{array}
\usepackage{tikz}

\newcommand\extrafootertext[1]{%
    \bgroup
    \renewcommand\thefootnote{\fnsymbol{footnote}}%
    \renewcommand\thempfootnote{\fnsymbol{mpfootnote}}%
    \footnotetext[0]{#1}%
    \egroup
}

\title{Benchmarking Transcriptomics Foundation Models for Perturbation Analysis : one PCA still rules them all}

\author{%
  Ihab Bendidi \\
  Valence Labs \\
  Ecole Normale Supérieure \\
  Paris, France \\
  \And
  Shawn Whitfield \\
  Valence Labs \\
  Montreal, Canada \\
  \And
  Kian Kenyon-Dean \\
  Recursion \\
  Toronto, Canada \\
  \AND
  Hanene Ben Yedder \\
  Valence Labs \\
  Montreal, Canada \\
  \And
  Yassir El Mesbahi \\
  Valence Labs \\
  Montreal, Canada \\
  \And
  Emmanuel Noutahi \\
  Valence Labs \\
  Montreal, Canada \\
  \And
  Alisandra K. Denton \\
  Valence Labs \\
  Montreal, Canada \\
}

\begin{document}

\maketitle

\begin{abstract}
  Understanding the relationships among genes, compounds, and their interactions in living organisms remains limited due to technological constraints and the complexity of biological data. Deep learning has shown promise in exploring these relationships using various data types. However, transcriptomics, which provides detailed insights into cellular states, is still underused due to its high noise levels and limited data availability. Recent advancements in transcriptomics sequencing provide new opportunities to uncover valuable insights, especially with the rise of many new foundation models for transcriptomics, yet no benchmark has been made to robustly evaluate the effectiveness of these rising models for perturbation analysis. This article presents a novel biologically motivated evaluation framework and a hierarchy of perturbation analysis tasks for comparing the performance of pretrained foundation models to each other and to more classical techniques of learning from transcriptomics data. We compile diverse public datasets from different sequencing techniques and cell lines to assess models performance. Our approach identifies scVI and PCA to be far better suited models for understanding biological perturbations in comparison to existing foundation models, especially in their application in real-world scenarios. 
\end{abstract}

\extrafootertext{Code and data :\href{https://github.com/valence-labs/Tx-Evaluation}{ https://github.com/valence-labs/Tx-Evaluation}}

\section{Introduction}



Living organisms are composed of countless molecular components that interact in complex and varied ways, the majority of which remain poorly understood. This gap in understanding is partly due to the limitations of current technological tools, which cannot fully measure or observe all the components of life across different conditions. Although recent advances \citep {stahl2016visualization, Dixit2016-tl, Chandrasekaran2023-pg} have allowed partial measurement and observation of key components under specific circumstances, our understanding of the relationships among genes, compounds, their functions, and their interactions remains limited \citep{Conesa2016-zm, Kharchenko2021-qg}. As a result, our view of these biological components is still partial and expensive to obtain, preventing a complete understanding of their behavior over time or across different states.

To better understand the function of cells and their components, one common approach is to perturb specific elements, such as genes, and observe the resulting effects on the cell and other genes. However, with over 20,000 protein-coding genes in the human genome and around \(10^{60}\) potential compounds in the chemical space \citep{Reymond2015-ae}, manually exploring these relationships is impractical. The complexity of these combinatorial possibilities necessitates the use of computational methods to narrow down the problem. Recent advancements in deep learning have shown promise in using existing data to uncover new relationships and functions. For example, deep learning models have been used to predict protein folding from their sequences \citep{Jumper2021}, understand binding dynamics \citep{corso2023diffdockdiffusionstepstwists, Evans2021-vm}, and identify novel biological relationships through analysis of large-scale microscopy imaging data of perturbed cells without use of prior knowledge \citep{kraus2024maskedautoencodersmicroscopyscalable}.

While various modalities of biological data have been successfully employed for perturbation analysis, transcriptomics stands out by offering a structured, detailed, and near-to-biological-mechanism view of cellular states compared to microscopy imaging of cells post-perturbation \citep{Williams2022-xx,Camunas-Soler2024-se}, while being faster and more cost-effective than proteomics. Despite this potential, transcriptomics has been underutilized, primarily due to low technological maturity and a scarcity of curated datasets linked to specific biological perturbations. However, recent developments in transcriptomics sequencing techniques and datasets focused on perturbations, such as those from \cite{Replogle2022-lx} using Perturb-Seq \citep{Dixit2016-tl, Adamson2016-my}, provide new opportunities to extract valuable insights similar to those achieved with the imaging modality \citep{Chandrasekaran2023-pg}. Pretrained foundation models for transcriptomics, trained on large-scale datasets using diverse techniques, offer a promising approach to harness these insights. Nevertheless, it remains unclear which models are most effective for perturbation analysis, as this was not their primary focus, and while exploratory efforts have been made \citep{Ahlmann-Eltze2024-yu}, there is no comprehensive benchmark  to evaluate their performance in this context. 

In this work, we tackle this challenge by developing a benchmark for biologically relevant perturbation tasks, through the curation of existing biological tasks, and the introduction of \textit{Structural Integrity} as a new evaluation task. Using public datasets from three sequencing techniques across various cell lines, we assess model performance on medium- to large-scale perturbation data. Our systematic comparison identifies scVI \citep{scvilopez2018deep} and PCA as more effective for analyzing biological perturbations compared to existing transcriptomics foundation models.

\section{Related Works}
\label{gen_inst}
\paragraph{Transcriptomics Data.} Measurements of gene expression levels provide valuable insights into cellular states, with various techniques applied in different contexts and formats. These include single cell RNA-Seq (scRNA-Seq) \citep{tang2009mrna} and bulk RNA-Seq \citep{Emrich2007-nr, mortazavi2008mapping}, which are commonly used to sequence gene expression levels, and spatial transcriptomics \citep{stahl2016visualization}, which is used to measure spatially resolved gene expression levels on histopathology slides. Perturbation is a critical method in biology where a targeted change is made to the abundance, state, or activity of a (macro) molecule, which helps disentangle causation from correlation and elucidate molecular mechanisms. Both chemical and genetic perturbations can be combined with gene expression assays. Drug-Seq \citep{ye2018druggable} measures gene expression changes resulting from chemical perturbations, while Perturb-Seq \citep{Dixit2016-tl, Adamson2016-my} uses CRISPR gene knockouts to measure changes in genetic perturbations. Similarly, L1000 \citep{subramanian2017next} is a hybridization-based approach that measures the expression of nearly 1,000 landmark genes, allowing both chemical and genetic screens. The data from the transcriptomics methods above is high dimensional and noisy, necessitating dimensionality reduction and noise-reduction techniques. Historically, analysis has focused on specific datasets or sequencing techniques, using methods such as PCA \citep{jolliffe2002principal, Bao2022-go} and autoencoders \citep{Chandrasekaran2023-pg,scvilopez2018deep}. However, with the rise of massive computational infrastructures, there has been a shift towards more advanced methods for learning from transcriptomics data.

\paragraph{Transcriptomics Foundation Models.} scVI  \citep{scvilopez2018deep}, a variational autoencoder tailored for single-cell RNA sequencing,  has been commonly used as the gold standard approach for transcriptomics analysis. It focuses on efficient dimensionality reduction and denoising, often serving as a robust baseline for comparison and has recently been adapted to support efficient transfer learning \citep{lotfollahi2022mapping}. However, the growing availability of transcriptomics data has led to the development of several foundation models trained on large-scale datasets \citep{Megill2020-qe}. Transformer-based models like Geneformer \citep{chen2023geneformer}, scGPT \citep{wang2023scgpt}, CellPLM \citep{chen2021single} and UCE \citep{Rosen2023-gg} are designed to capture complex gene expression patterns across different contexts, and they claim to outperform scVI on various downstream tasks. Geneformer \citep{chen2023geneformer} uses a transformer architecture with universal gene embeddings to support diverse tasks like cell type classification and perturbation prediction, and employs a rank-based tokenization to prioritize genes with regulatory significance. scGPT \citep{wang2023scgpt} treats single-cell RNA-seq data as sequences of gene tokens and uses a decoder-only transformer (GPT architecture) to model sequential dependencies, addressing data sparsity. CellPLM \citep{chen2021single} introduces a Gaussian mixture latent space, enabling the model to capture both intra- and intercellular information using masked language modeling to learn gene and cell relationships. Universal Cell Embeddings (UCE) \citep{Rosen2023-gg} uses protein embeddings as tokens for each gene, providing a unified representation space for cross-dataset comparisons. Overall, scVI, Geneformer, scGPT, CellPLM, and UCE are widely used due to their proclaimed diverse capabilities and performance across various transcriptomic analyses.

\paragraph{Perturbation Analysis and Benchmarks.}  The assays of genetic perturbations described above are designed to capture the effects of targeted molecular changes that are both multitudinous, and often subtle. Existing computational models \citep{chen2021single, wang2023scgpt, Hrovatin2023} and benchmarks \citep{Boiarsky2023-df, Alsabbagh2023-ib, Kedzierska2023-ll} alike often focus on broad biological tasks like cell type classification, clustering, and batch integration. Limited studies include the occasional perturbation-oriented metric (e.g. metrics on post-perturbation expression in both scEval \citep{Liu2023-pz} and scGPT \citep{wang2023scgpt}). Beyond just transcriptomics, recent efforts, such as those to standardize perturbation metrics on mean average precision \citep{Kalinin2024-ri} or establish application-ready pipelines to benchmark and generate biological maps from perturbational data, EFAAR \citep{Celik2022} have made strides in scoring especially, but not only, microscopy-based perturbation studies. For instance such methods proved useful in evaluating representations of self-supervised models like Masked Autoencoders \citep{kraus2024maskedautoencodersmicroscopyscalable} and Set-DINO \citep{yao2024weaklysupervisedsetconsistencylearning} for microscopy-based perturbation studies. However, a comprehensive benchmark that unifies all perturbation-related evaluation tasks across the Transcriptomics modality is still lacking, despite its importance for advancing the predictive capabilities of models in understanding cellular responses and therapeutic development.

\section{Perturbation Hierarchy of Evaluation Metrics}
\label{benchmark}

Understanding how biological systems respond to interventions—such as gene knockouts or drug treatments—requires models that can accurately recognize and predict these effects. However, evaluating these models is complex because a model may excel in one area but struggle in another, making it difficult to assess its overall usefulness. To address this challenge, we propose a structured hierarchy of evaluation metrics to assess models in a clear and systematic way. By progressing through this hierarchy in order, a model demonstrates both technical competence and real-world applicability for perturbation analysis. The structure ensures that fundamental criteria are met first before moving on to more specific or complex tasks, allowing for a more holistic evaluation. Detailed mathematical descriptions of each metric are provided in the Appendix.

\subsection{Data Integration and Batch Effect Reduction} 

In biological experiments, data often come from different batches. A \textit{batch} is a sets of samples processed at different times or under slightly different conditions. These batch differences can introduce artificial variations known as \textit{batch effects}, which can obscure true perturbation or treatment response signals. For perturbation analysis, it is crucial that a model can integrate data from multiple batches seamlessly, ensuring that comparisons between samples reflect real biological differences, not technical artifacts. We use the Integration Local Inverse Simpson's Index (iLISI) metric \citep{korsunsky2019fast} to measure how well a model reduces batch effects. This metric assesses how well samples from different batches are mixed within the model's representation space. If, in the neighborhood of any given sample, there's a good mix of samples from all batches, it suggests that the model has effectively minimized batch effects. 

To compute the iLISI score, we first define the conditional probability \( p_{ij} \) of sample \( i \) selecting sample \( j \) as a neighbor, with \( d_{ij} \) being the distance between samples \( i \) and \( j \)  :

\[
p_{ij} = \frac{\exp(-\beta_i d_{ij})}{\sum_{l \in \mathcal{N}_i} \exp(-\beta_i d_{il})}
\]

Where \( \beta_i \) is a scaling parameter adjusted such that the entropy \( H(P_i) = \log(k) \) of the distribution \( P_i = \{ p_{ij} \} \) ensures the number of nearest neighbors matches the target number of neighbors $k$, and \( \mathcal{N}_i \) is the set of \( k \) nearest neighbors of sample \( i \). 

With \( n \) being the total number of samples, \( C \) the set of all possible categories (batch labels), and \( l_j \) the label (e.g., batch category) of neighbor \( j \), the iLISI score is then calculated as:

\[
\text{iLISI} = \frac{1}{n} \sum_{i=1}^{n} \left( \sum_{c \in C} \left( \sum_{\substack{j \in \mathcal{N}_i \\ l_j = c}} p_{ij} \right)^2 \right)^{-1}
\]

\subsection{Latent Space Linear Separability of Known Perturbations} 

After ensuring that batch effects are minimized, the next step is to verify that the model can distinguish between different perturbations. In other words, the model's internal "map" (or latent space) should reflect the biological differences caused by various perturbations \textbf{in its global structure}. This capability is crucial for mapping how different interventions affect biological systems. We assess this by checking if samples subjected to different perturbations can be separated using a simple linear classifier for linear probing. Specifically, we add a straightforward linear layer  on top of the model's representations to classify samples based on their perturbations. We then evaluate the classifier's performance on the same perturbations but on new biological batches of data that the model hasn't seen before. This step ensures that the model is not just memorizing noise patterns of the training data but can generalize to new, unseen data.

\subsection{Perturbation Consistency} 

Recognizing different perturbations is important, but it is equally critical that the model represents each perturbation consistently across various samples and batches. This consistency ensures the model's robustness, making its representations reliable and not influenced by noise or outliers. To measure this, we use the \emph{Perturbation Consistency} metric introduced in \citep{Celik2022}.

For each gene perturbation \( g \), we calculate the cosine similarity between all pairs of its embeddings from different samples and batches. The average of these similarities gives the perturbation similarity score for that perturbation. Formally, let \( \mathbf{x}_{g,i} \) be the embedding vector for the \( i \)-th sample of perturbation \( g \), and \( n_g \) be the number of samples for \( g \). A per-perturbation similarity score \( \text{avgsim}_g \) is computed as:

\[
\text{avgsim}_g = \frac{1}{n_g^2} \sum_{i=1}^{n_g} \sum_{j=1}^{n_g} \frac{\langle \mathbf{x}_{g,i}, \mathbf{x}_{g,j} \rangle}{\|\mathbf{x}_{g,i}\| \, \|\mathbf{x}_{g,j}\|}
\]

This similarity score is then compared to a null distribution generated from unexpressed genes or genes that are inactive in the dataset. Unexpressed genes are selected based on their consistently low expression levels, with at least 1000 unexpressed genes required for meaningful results. For each unexpressed gene \( g_k' \), \( k = 1, \dots, K \), we compute their average cosine similarity \( \text{avgsim}_{g_k'} \) in the same way. Using a permutation test, we assess whether the observed similarity for perturbation \( g \) is significantly higher than what would occur by chance. The consistency $p$-value for each gene \( g \) is given by:

\[
p_g = \frac{\max\left\{ \#\left( \text{avgsim}_{g_k'} \leq \text{avgsim}_g \right), 1 \right\}}{K}
\]

Gene perturbations that achieved a consistency $p$-value $< 0.05$ are considered significant. The Perturbation Consistency metric reports the fraction of such significant genes compared to all gene perturbations. A high perturbation consistency score indicates that the model consistently recognizes the effect of a perturbation across different batches and experiments. Conversely, a low score suggests that the model may not fully capture the perturbation's effect, potentially classifying it correctly in other metrics simply because of outlier behavior or similarity to other cases.

\subsection{Latent Space Direct Organization} 
Beyond perturbation consistency, we aim for the model's latent space to self-organize without extra help or training. A linear classifier can still perform well even if clusters are fuzzy or poorly separated, as it focuses on global structure of the latent space. Thus, a high linear separability score does not necessarily mean the latent space is locally well organized. Once the global structure is validated, we focus on the local structure of the perturbation space, useful for tasks like retrieval. We achieve that by directly evaluating how well-defined and locally organized perturbation clusters are, especially in unseen data. Such a local organization is crucial for the model to be useful in practical, exploratory settings where we would like to retrieve and compare cells from different conditions but similar perturbations. To assess this, we apply a k-Nearest Neighbors (kNN) algorithm using two different datasets with the same perturbations (a query set and reference set),  but with no overlap of biological batches.
For each sample in the test set, we identify its closest neighbors in the latent space of the reference set. We compute the kNN accuracy using the samples from the query batches that correspond to the same perturbation as their (\(k = \lfloor \sqrt{n} \rfloor\), $n$ being the total number of samples in reference set) closest neighbors from the reference set of batches.

\subsection{Zero-Shot Retrieval of Known Biological Relationships}

A strong model should be able to capture biological relationships between genes and even discover new ones, without being explicitly trained to find these connections. This ability is essential for generating new insights and validating the model’s biological relevance. To evaluate this, we use the known relationships retrieval metric introduced by \citep{Celik2022}, which assesses how well gene-to-gene distances in the model’s latent space can retrieve known relationships from curated gene/protein interaction databases like CORUM \citep{Giurgiu2019-jt}, HuMAP \citep{Drew2021-aa}, Reactome, SIGNOR, and StringDB \citep{Celik2022}. This metric evaluates the model's ability to discover relationships that exist but were not explicitly provided during training, highlighting its potential in exploratory settings where unknown relationships are sought. 

We compute our metric by first calculating pairwise cosine similarities between the aggregated perturbation embeddings of all perturbed genes. We exclude self-links (similarities of a gene with itself) since their similarity is always one, which would distort the results. Next, we selected the strongest relationships with cosine similarities falling below the 5th percentile or above the 95th percentile as ``predicted links''. We focus on recall as a metric because we are primarily interested in how many true relationships the model can identify from its top predictions. Precision is not applicable for evaluation as we lack ground truth of the other relationships' correctness. We then compute the recall by comparing these predicted links with known gene-gene relationships from the benchmark databases. For each database, recall is defined as the proportion of true relationships (i.e., known links) retrieved by the model, relative to the total number of known gene-gene relationships in that database that are also present in the perturbation dataset. This adjustment ensures fairness when comparing datasets with different numbers of genes. The recall values are then multiplied by 100 to express them as percentages.

\subsection{Linear Interpretability of Latent Space}

A key goal is to interpret a good model's internal representations in terms of actual gene activity. By decoding the latent space back into gene expression profiles, we can better understand the biological basis for the model’s predictions and discoveries. To fairly evaluate how accurately the latent embeddings can be decoded into gene expression data, we train a simple linear multilayer perceptron (MLP) on top of a frozen model to map the latent space back to gene expression counts. We assess the quality of this reconstruction using Spearman correlation between true and reconstructed expressions. This metric shows strong correlation with MSE, MAE and Pearson correlation (See Appendix). These metrics provide insight into how well the latent space reflects true gene expression patterns.

We also introduce the notion of \emph{Structural Integrity}, a novel metric designed to assess how well a model's embeddings preserve the relationship between control and perturbation conditions within each biological batch in the gene activity dimension. This is crucial for utilizing reconstructed gene expression profiles to study gene expression changes under different conditions. Specifically, we first center the log-transformed gene expression profiles by subtracting the corresponding control (predicted vs. actual) within each biological batch from the perturbed gene expression profiles. This centering accounts for batch-specific variability and focuses on the perturbation effects. The structural distance is then computed for each batch $b$ as the Frobenius norm of the difference between the centered predicted and centered actual gene expression matrices:

\begin{equation}
    \text{Structural Distance} = \frac{1}{B} \sum_{b=1}^{B} \frac{1}{n_b} \left\| \tilde{Y}_{\text{pred}}^{(b)} - \tilde{Y}_{\text{actual}}^{(b)} \right\|_F,
\end{equation}

where $B$ is the total number of batches, and \( n_b \) is the number of samples in batch \( b \). $\tilde{Y}_{\text{pred}}^{(b)}$ and $\tilde{Y}_{\text{actual}}^{(b)}$ are the centered predicted and actual gene expression matrices for batch $b$, respectively, and $\|\cdot\|_F$ denotes the Frobenius norm. The theoretical upper bound for the structural distance, given the number of unique measured genes \( g \) and assuming the gene library size is \( M \), is:

\begin{equation}
    \text{Structural Distance}_{max} = \frac{2}{B} \sum_{b=1}^{B} M \sqrt{ n_b \times g } \approx \frac{2}{B} \sum_{b=1}^{B} \frac{1}{n_b} \left\| \tilde{Y}_{\text{actual}}^{(b)} \right\|_F,
\end{equation}

Then, the \emph{Structural Integrity} is computed as:

\begin{equation}
    \text{Structural Integrity} = 1 - \frac{\text{Structural Distance}}{\text{Structural Distance}_{max}}.
\end{equation}

Higher values of Structural Integrity indicate better preservation of the structural relationships in gene expression data. This metric provides an assessment of how well the model captures the overall structure of gene expression changes while accounting for batch-specific variability.

\section{Benchmark Experimental Setup}
\label{setup}


\begin{table}[bt]
  \caption{Summary of datasets used for model evaluation}
  \label{tab:datasets}
  \centering
  \small 
  \begin{tabularx}{\textwidth}{
    >{\raggedright\arraybackslash}X 
    >{\centering\arraybackslash}X   
    >{\centering\arraybackslash}X   
    >{\centering\arraybackslash}X   
  }
    \toprule
    \textbf{Dataset}            & \textbf{\citet{Replogle2022-lx}} & \textbf{L1000 CRISPR} & \textbf{\citet{Joung2023-nr}} \\
    \midrule
    \textbf{Sample Type}        & Single-cell            & Bulk                  & Single-cell                  \\
    \textbf{Number of Samples}  & 1.98M                  & 443,365               & 1.38M                        \\
    \textbf{Control Samples}    & 75,328                 & 87,565                & 173,211                      \\
    \textbf{Perturbations}      & 9,866 Knockouts        & 5,157 Knockouts       & 1,762 Overexpression         \\
    \textbf{Cell Lines}         & 1 (K562)               & 31                    & 1 (H1 hESC)                  \\
    \textbf{Measured Genes}     & 8,248                  & 1K (+11K Inferred) & 37,528                      \\
    \textbf{Batches/Experiments}& 267                    & 1,188                 & 2                            \\
    \textbf{Usage}              & Train/Eval             & Train/Eval            & Train                        \\
    \bottomrule
  \end{tabularx}
  \vskip -0.1in
\end{table}

We benchmark several advanced models for zero-shot single-cell analysis, such as scGPT \citep{wang2023scgpt}, Geneformer \citep{chen2023geneformer}, CellPLM \citep{chen2021single}, and Universal Cell Embeddings (UCE) \citep{Rosen2023-gg}, against simpler baselines like PCA and scVI \citep{scvilopez2018deep}. These models, built on transformer architectures \citep{vaswani2023attentionneed} and trained on massive datasets \citep{Megill2020-qe}, differ in how they handle gene expression data. For instance, scGPT and Geneformer focus on gene-gene relationships, while CellPLM uses a latent space to capture both intra- and intercellular interactions. UCE integrates gene and protein data for a more universal representation. Further implementation details are provided in the Appendix.

We apply several post-processing approaches to the embeddings from each model to determine which works best for each model and task. A baseline method for aligning perturbation units involves using control units in each batch to center and scale features. Additionally, we implement Typical Variation Normalization (TVN) \citep{Ando2017-ar}, which aligns not only the first-order statistics but also the covariance structures of the data. We also compare using raw embeddings without any post-processing. Since our evaluation emphasizes practical use cases related to perturbations, and because different models respond differently to various post-processing methods, we select the best-performing post-processing for each model in each task. It is important to note that the optimal post-processing method can vary, not only between models but even across different tasks for the same model. In a real-world scenario, users would choose the post-processing technique that yields the best results for their specific application. Therefore, we report the best post-processing method for each model and task in the benchmark. Full results, including scores for all models, tasks, and post-processing approaches, are available in the appendix.

\begin{table}[tb]
\caption{Comparison of model performance on the Replogle and L1000 datasets across the hierarchy of tasks. Higher is better for all metrics. Each score is the average of 5 different runs. Std and detailed scores are available in Appendix.}
\label{tab:t_overall_results}
    \centering
    \resizebox{\textwidth}{!}{%
    \begin{tabular}{c l c c c c c c c c c}
        \toprule
        \textbf{Task} & \textbf{Metric} & \textbf{Rand. Labels} & \textbf{Geneformer} & \textbf{UCE} & \textbf{cellPLM} & \textbf{scGPT} & \textbf{scGPT finetuned} & \textbf{Transfer scVI} & \textbf{scVI} & \textbf{PCA} \\
        \midrule
         & & \multicolumn{9}{c}{\textbf{\citep{Replogle2022-lx}}} \\
        \cmidrule(lr){3-11}
        \multirow{1}{*}{\textbf{1}} & \textbf{iLISI} & \textbf{1.000} & \underline{0.979} & 0.931 & 0.839 & 0.941 & 0.954 & 0.881 & 0.874 & 0.919 \\
        \cmidrule(lr){2-11}
        \multirow{2}{*}{\textbf{2}} & \textbf{Top5 lin.} & 0.005 & 0.005 & 0.008 & 0.009 & 0.011 & 0.005 & 0.010 & \underline{0.032} & \textbf{0.058} \\
        & \textbf{Top1 lin.} & 0.001 & 0.002 & 0.003 & 0.003 & 0.004 & 0.002 & 0.004 & \underline{0.016} & \textbf{0.033} \\
        \cmidrule(lr){2-11}
        \multirow{1}{*}{\textbf{3}} & \textbf{Pert Cons.} & 0.051 & 0.063 & 0.105 & 0.111 & 0.115 & 0.092 & \underline{0.118} & 0.102 & \textbf{0.119} \\
        \cmidrule(lr){2-11}
        \multirow{2}{*}{\textbf{4}} & \textbf{Top5 knn} & 0.054 & 0.054 & 0.055 & \underline{0.056} & \underline{0.056} & 0.054 & 0.055 & \textbf{0.061} & \textbf{0.061} \\
        & \textbf{Top1 knn} & 0.053 & 0.053 & 0.053 & 0.053 & 0.053 & \textbf{0.054} & 0.053 & \textbf{0.054} & \textbf{0.054} \\
        \cmidrule(lr){2-11}
        \multirow{5}{*}{\textbf{5}} & \textbf{CORUM} & 0.107 & 0.149 & 0.366 & 0.284 & 0.389 & 0.166 & 0.443 & \textbf{0.503} & \underline{0.450} \\
        & \textbf{HuMAP} & 0.102 & 0.135 & 0.253 & 0.272 & 0.323 & 0.149 & 0.346 & \textbf{0.388} & \underline{0.359} \\
        & \textbf{Reactome} & 0.090 & 0.117 & 0.170 & 0.144 & 0.166 & 0.116 & 0.174 & \textbf{0.225} & \underline{0.213} \\
        & \textbf{SIGNOR} & 0.097 & 0.108 & 0.128 & 0.121 & 0.123 & 0.111 & 0.123 & \underline{0.147} & \textbf{0.150} \\
        & \textbf{StringDB} & 0.102 & 0.148 & 0.329 & 0.305 & 0.402 & 0.170 & 0.428 & \textbf{0.497} & \underline{0.475} \\
        \cmidrule(lr){2-11}
        \multirow{2}{*}{\textbf{6}} & \textbf{Spear. Corr} & 0.000 & 0.001 & 0.172 & 0.169 & 0.180 & 0.131 & \underline{0.218} & \textbf{0.265} & 0.215 \\
        & \textbf{Struct. Int.} & 0.525 & 0.528 & 0.538 & 0.538 & 0.540 & 0.532 & 0.548 & \textbf{0.551} & \underline{0.550} \\
        \midrule
         & & \multicolumn{9}{c}{\textbf{L1000 Assay}} \\
        \cmidrule(lr){3-11}
        \multirow{1}{*}{\textbf{1}} & \textbf{iLISI} & \textbf{1.000} & 0.691 & \underline{0.699} & 0.100 & 0.660 & 0.682 & 0.467 & 0.408 & 0.398 \\
        \cmidrule(lr){2-11}
        \multirow{2}{*}{\textbf{2}} & \textbf{Top5 lin.} & 0.017 & 0.022 & 0.022 & 0.022 & 0.022 & 0.022 & 0.026 & \underline{0.044} & \textbf{0.053} \\
        & \textbf{Top1 lin.} & 0.007 & 0.008 & 0.008 & 0.008 & 0.008 & 0.008 & 0.010 & \underline{0.021} & \textbf{0.027} \\
        \cmidrule(lr){2-11}
        \multirow{1}{*}{\textbf{3}} & \textbf{Pert Cons.} & 0.047 & 0.059 & 0.058 & 0.068 & 0.056 & 0.058 & 0.064 & \textbf{0.082} & \underline{0.073} \\
        \cmidrule(lr){2-11}
        \multirow{2}{*}{\textbf{4}} & \textbf{Top5 knn} & 0.261 & 0.262 & 0.262 & \underline{0.270} & 0.263 & 0.262 & 0.269 & 0.268 & \textbf{0.271} \\
        & \textbf{Top1 knn} & 0.256 & 0.256 & 0.256 & \textbf{0.258} & 0.256 & 0.256 & 0.256 & 0.256 & 0.256 \\
        \cmidrule(lr){2-11}
        \multirow{5}{*}{\textbf{5}} & \textbf{CORUM} & 0.111 & 0.141 & 0.163 & 0.111 & 0.160 & 0.117 & \textbf{0.203} & \underline{0.190} & 0.174 \\
        & \textbf{HuMAP} & 0.113 & 0.124 & 0.121 & 0.116 & 0.147 & 0.107 & \textbf{0.173} & \underline{0.170} & 0.153 \\
        & \textbf{Reactome} & 0.109 & 0.128 & 0.141 & 0.124 & 0.124 & 0.119 & 0.141 & \textbf{0.149} & \underline{0.144} \\
        & \textbf{SIGNOR} & 0.120 & 0.145 & \textbf{0.204} & 0.114 & 0.128 & 0.104 & \underline{0.164} & 0.134 & 0.139 \\
        & \textbf{StringDB} & 0.109 & 0.150 & 0.146 & 0.147 & 0.137 & 0.127 & \textbf{0.165} & \underline{0.162} & 0.161 \\
        \cmidrule(lr){2-11}
        \multirow{2}{*}{\textbf{6}} & \textbf{Spear. Corr} & 0.002 & 0.748 & 0.253 & 0.811 & 0.833 & 0.757 & \underline{0.903} & \textbf{0.928} & 0.882 \\
        & \textbf{Struct. Int} & 0.938 & 0.955 & 0.942 & 0.960 & 0.962 & 0.955 & 0.970 & \textbf{0.977} & \underline{0.974} \\
        \bottomrule
    \end{tabular}
    } 
    \vskip -0.1in
\end{table}


We use three primary open source datasets (Table \ref{tab:datasets}) : the single-cell gene knockout dataset \citep{Replogle2022-lx}, the bulk L1000 CRISPR assay \citep{Subramanian2017-rx}, and the single cell gene overexpression dataset \citep{Joung2023-nr}, each capturing different aspects of gene perturbations.  Detailed dataset characteristics can be found in Table \ref{tab:datasets}. \citep{Joung2023-nr} dataset is used mainly for training models on gene overexpression, before evaluating the trained model on gene knockout on different cell lines. For consistency, we preprocess all datasets to use the same metadata and structure, and retain samples with total raw counts above 1,000. For evaluation, we apply a 70\% train/test split for linear probing and kNN, ensuring distinct batches in the test set but retaining the same perturbations. Reconstruction tasks use different batches and perturbations in the test set, and all data is used for perturbation consistency and known relationship recall tasks.

\section{Results}
\label{results}

\subsection{Current Foundation models do not generalize to perturbation related tasks}


%

The results in Table \ref{tab:t_overall_results} demonstrate that current foundation models do not generalize well to perturbation-related tasks compared to simpler approaches like PCA and scVI. PCA, which is applied on raw gene counts, and scVI, trained from scratch on the same dataset, consistently outperform foundation models across most tasks, except for batch effect reduction (Task 1). Notably, scVI achieves strong performance in both scenarios: when trained directly on the evaluation dataset (scVI) and when used in a zero-shot transfer learning context (Transfer scVI), in which it is pre-trained on a different cell line, perturbation type, and sequencing technique \citep{Joung2023-nr} before being evaluated on \citep{Replogle2022-lx} and L1000 data. Transfer scVI ranks third overall, after PCA and scVI, highlighting its robustness in handling strong out-of-distribution scenarios. Interestingly, Transfer scVI consistently surpasses scVI for batch effect reduction, as this metric is easily optimized by capturing higher levels of noise, which is the case for transfer learning zero-shot applications. Additionally, PCA shows higher structural integrity than Transfer scVI for both the L1000 assay and the \citep{Replogle2022-lx} dataset, while Transfer scVI achieves a higher Spearman correlation for reconstructing expression counts. This can be explained by the fact that PCA preserve the original data structure, while Transfer scVI conserves \citep{Joung2023-nr} original data structure, even if trained to reconstruct \citep{Replogle2022-lx} and L1000 dataset. Such cases suggest that structural integrity and Spearman correlation capture different aspects of model performance and complement each other. 

Foundation models such as Geneformer and scGPT show competitive performance only in batch effect reduction, where random embeddings achieve near-optimal results, but they struggle across more biologically meaningful tasks. This suggests that their training objectives, which likely focus on reducing batch effects, are insufficient for capturing nuanced biological insights required for perturbation tasks. Interestingly, finetuning scGPT on the same evaluation data improves its performance on batch effect reduction but has little to no effect on linear separability of perturbations (Task 2) and dramatically reduces performance on zero-shot recall of known biological relationships (Task 5). This hints that its learning objective is not adapted to learning relevant representations of perturbation biology even when trained on its evaluation data. The overall results indicate that while foundation models can be tuned for specific technical metrics, they do not yet effectively generalize to biologically complex tasks like perturbation analysis, where scVI and PCA remain more reliable.

\subsection{Gene distribution matters and is dataset-dependent}

\begin{figure}[tb]
\begin{center}
\centerline{\includegraphics[width=\textwidth]{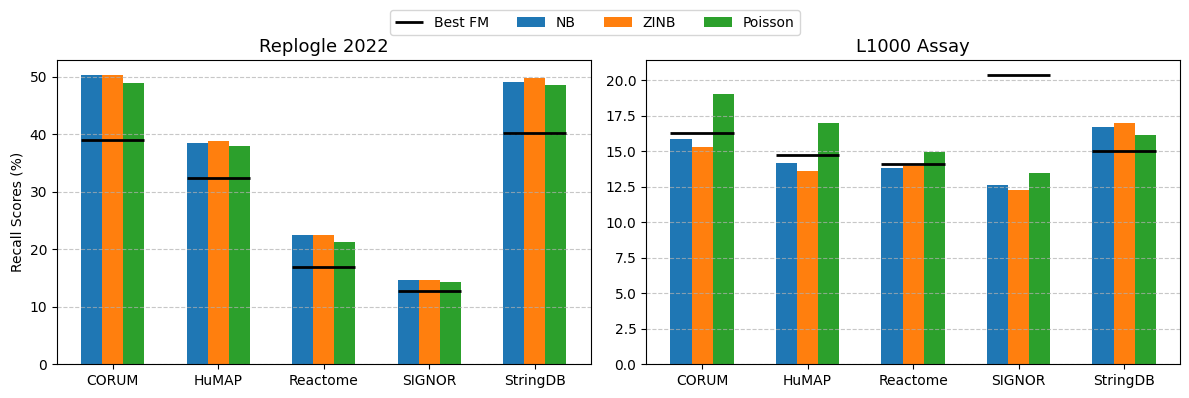}}
    \caption{Known biological relationship recall scores for \citep{Replogle2022-lx} and L1000 Assay for scVI, trained using different gene distributions. Different datasets and sequencing approaches benefit from different gene distributions.}
    \label{fig:gene_dist}
\end{center}
\vskip -0.3in
\end{figure}

We trained scVI using three different distributions: Zero-Inflated Negative Binomial (ZINB), Negative Binomial (NB), and Poisson, and evaluated their performance on the \cite{Replogle2022-lx} and L1000 Assay datasets, reporting results on zero-shot recall of known biological relationships (Task 5) in Figure \ref{fig:gene_dist}. For \cite{Replogle2022-lx}, the ZINB distribution, originally introduced for scVI \citep{scvilopez2018deep}, achieves the best performance, slightly outperforming NB, while Poisson performs slightly worse. However, scVI's performance remains stable across all gene distributions on this dataset, consistently surpassing the best-performing foundation models for this task. In contrast, for the L1000 dataset, Poisson distribution yields the highest performance, substantially outperforming ZINB and NB. This discrepancy can be attributed to the nature of the L1000 assay, where gene expression levels are measured as continuous values rather than discrete counts typically obtained from sequencing. As a result, the Poisson distribution, which can accommodate continuous data under certain conditions, yields better performance compared to ZINB and NB distributions designed for overdispersed count data. Thus, gene distribution plays a crucial role in determining scVI’s overall performance, particularly for different assay techniques like L1000 and the associated datasets.


\subsection{scVI robustly learns from small-scale single cell perturbation data}

\begin{figure}[tb]
\begin{center}
\centerline{\includegraphics[width=\textwidth]{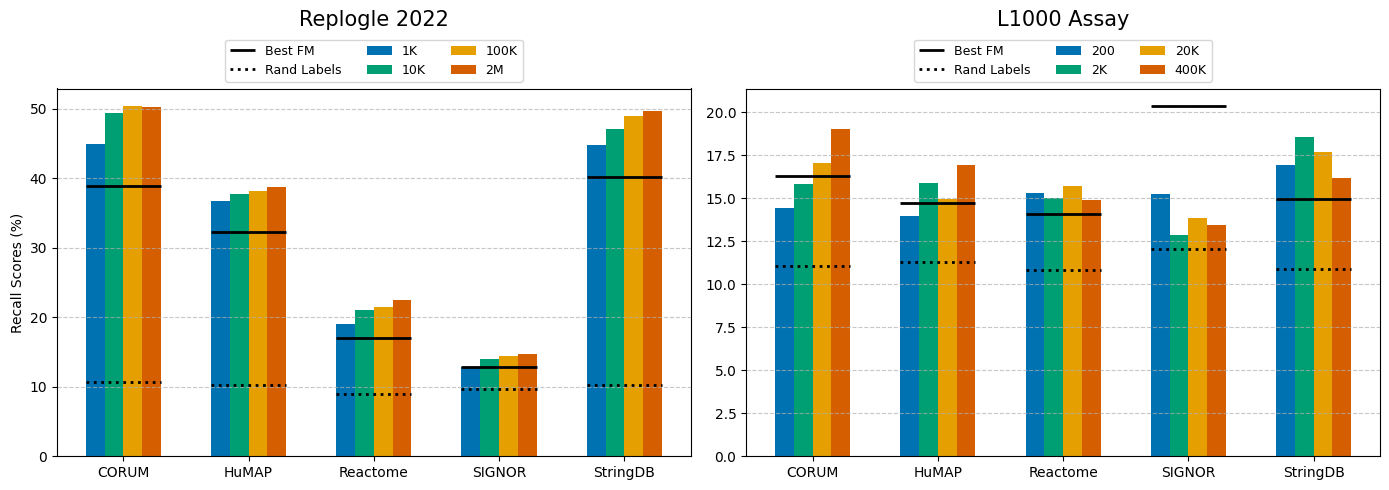}}
    \caption{Training scVI on different amounts of samples, using distinct batches and perturbations, before evaluating known biological relationship retrieval. scVI is robust to very low data regime, and shows strong data scaling laws on single cell data.}
    \label{fig:downscaling_scvi}
\end{center}
\vskip -0.3in
\end{figure}

We trained scVI on different subsets of the Replogle and L1000 datasets, varying the number of samples used for training, and evaluated the models on the full datasets for the zero-shot recall of known relationships (Task 5). The subsets were created by selecting batches that constituted the target ratio, ensuring that most batches and perturbations present in the evaluation data were unseen during training to mimic a real-world scenario where data is acquired incrementally. We used the ZINB gene distribution for \cite{Replogle2022-lx}, and used the Poisson distribution for the L1000 assay. The results in Figure \ref{fig:downscaling_scvi} show that scVI consistently learns useful representations for zero-shot biological tasks, even when trained on minimal data, with performance remaining above that of foundation models even in the lowest data regime (1000 samples). Moreover, for \cite{Replogle2022-lx}, scVI exhibits clear scaling behavior, with zero-shot performance improving as more data is included, underscoring its robustness for single-cell perturbation analysis. In contrast, the L1000 assay does not show the same scaling pattern, with scVI performance varying unpredictably as data size changes, except for the CORUM task. This lack of scaling pattern on L1000 assay may be due to the noise and uncertainty inherent in the assay \citep{subramanian2017next}, as well as the fact that scVI’s gene distribution assumptions, optimized for single-cell data, may not be well-suited for bulk readouts like L1000. Nonetheless, scVI consistently performs above random for L1000, even when trained on only 200 samples, demonstrating its robustness under less ideal conditions.

\section{Conclusion}

This study presents a biologically motivated benchmarking tool for evaluating model performance on perturbation relevant tasks, while introducing \emph{Structural Integrity} as a new metric for assessing gene activity structure preservation. Through systematic comparisons using public datasets across different sequencing techniques and cell lines, we uncover using this set of tasks that simpler models like scVI and PCA outperform current transcriptomics foundation models in most perturbation-related tasks. Specifically, scVI exhibits strong performance both when trained directly on evaluation datasets and in zero-shot transfer learning scenarios, highlighting its robustness and adaptability to diverse conditions. scVI also maintains robust performance even with minimal training data, and scales its performance with more training data in single-cell perturbation contexts. Our findings indicate that foundation models, while effective in batch effect reduction, struggle with more complex biological tasks, suggesting that their training objectives may not fully capture the nuances required for perturbation analysis. Additionally, we show that gene distribution assumptions significantly impact model performance and are highly dataset-dependent. These results indicate the need for developing models with biologically tailored objectives, and we offer a curated evaluation set designed to assess performance across a wider range of biologically relevant tasks, thereby measuring the utility of future models in uncovering complex biological relationships.

\newpage
\bibliography{iclr2024_conference}
\bibliographystyle{iclr2024_conference}

\newpage
\appendix
\section{Transcriptomics Foundation Models}


\subsection{Universal Cell Embeddings (UCE) Model}
\label{sec:uce_model}

The Universal Cell Embeddings (UCE) model \citep{Rosen2023-gg} is a foundational approach designed to generate a universal latent space that captures the biological variation of cells from multiple species and tissues. It leverages single-cell RNA sequencing (scRNA-seq) data to create a shared representation space that enables zero-shot inference for unseen datasets without further training.

UCE takes as input the gene expression count matrix of single cells and maps each cell into a unified embedding space through a series of transformations and self-supervised training. The model can handle cells from different species and datasets, addressing batch effects and variations in gene expression measurements. It achieves this by encoding gene information into a numerical representation using a pre-trained protein language model and subsequently applying a transformer architecture.

The input to the UCE model consists of two key elements:
\begin{itemize}
    \item \textbf{Gene expression counts} ($x_i \in \mathbb{R}^{K_i}$): The expression counts of $K_i$ genes in cell $c_i$. This can vary across cells and datasets due to differing gene availability.
    \item \textbf{Protein embeddings} ($p_g \in \mathbb{R}^{d_p}$): Each gene $g$ is represented by a protein embedding obtained using the ESM2 model \citep{Lin2022-px}, which takes an amino acid sequence as input and outputs a $d_p$-dimensional embedding vector.
\end{itemize}

The model represents each cell by constructing a "cell sentence" $S_i$, which includes tokens representing the expressed genes and their protein embeddings. The set of expressed genes for each cell is denoted as $G_i^+$, while the non-expressed genes are represented as $G_i^-$. The expressed genes are sampled with replacement based on their expression levels, as defined by the probability:
\begin{equation}
    P(g \mid c_i) = \frac{\log(x_i^g)}{\sum_{g' \in G_i^+} \log(x_i^{g'})},
\end{equation}
where $x_i^g$ is the expression count of gene $g$ in cell $c_i$.

The sampled genes are organized by chromosome, grouped together using special start and end tokens for each chromosome, and then concatenated to form the cell sentence $S_i$. The cell sentence $S_i$ is then passed into a transformer architecture.

UCE uses a transformer network with $n_{layers}$ layers and $n_{heads}$ attention heads per layer. Each gene's protein embedding is reduced to a lower-dimensional space using a multi-layer perceptron (MLP) before being processed by the transformer. A special classification token (CLS) is added to the beginning of the cell sentence, and its final output after passing through the transformer represents the cell embedding $h_i^{\text{cell}}$.

The UCE model is trained in a self-supervised manner using masked gene prediction. During training, a random subset of the expressed genes $G_i^+$ is masked, forming a masked set $G_i^{M+} \subset G_i^+$. The model predicts the expression status of both masked and non-expressed genes, formulated as a binary classification problem using the following loss function:
\begin{equation}
    L = -\frac{1}{N} \sum_{i=1}^{N} \frac{1}{N_{loss}} \sum_{j=1}^{N_{loss}} \left[y_{ij} \log(p(y_{ij})) + (1 - y_{ij}) \log(1 - p(y_{ij}))\right],
\end{equation}
where $y_{ij}$ is the true binary label indicating if gene $j$ is expressed in cell $c_i$, $N_{loss}$ the number of genes being predicted, and $p(y_{ij})$ is the predicted probability of gene expression.

The resulting UCE model can directly embed new datasets into the same latent space without any additional fine-tuning or training. This zero-shot capability enables UCE to generalize to new biological contexts and datasets, maintaining consistency across species and tissues.

UCE is implemented using the PyTorch library and consists of 33 transformer layers with 650 million parameters. The model was trained on over 300 datasets from 8 species, encompassing 36 million cells. Training was performed on 24 A100 GPUs for 40 days, using the CellXGene corpus \citep{Megill2020-qe} as the primary data source.

\subsection{scGPT Model}
\label{sec:scGPT_model}

The scGPT model \citep{wang2023scgpt} is a foundation model for single-cell multi-omics based on a generative pretraining approach using the transformer architecture. The core idea behind scGPT is to leverage the self-attention mechanism to capture complex relationships between genes and cells by learning meaningful embeddings. The model is pretrained on a large-scale single-cell RNA sequencing (scRNA-seq) dataset containing over 33 million human cells across various tissues and conditions.

scGPT consists of two main training stages: a general-purpose pretraining phase and a fine-tuning phase. In the pretraining stage, the model is trained in a self-supervised manner using a masked language modeling objective. This involves randomly masking out a portion of gene expression values and training the model to predict these masked values based on the remaining genes. To adapt the transformer architecture to non-sequential omics data, a specialized attention mask is introduced in the transformer’s multi-head attention blocks. The attention mechanism enables the model to attend to gene–gene and cell–gene interactions simultaneously, allowing for effective learning of gene expression patterns.

The input to the model consists of gene tokens, expression values, and condition tokens (e.g., batch, modality, or perturbation conditions). Each input token is embedded into a high-dimensional vector space, and the embeddings are combined through element-wise addition. The embeddings are then processed through multiple layers of transformer blocks, each consisting of a masked multi-head attention layer followed by a feed-forward layer. During the pretraining phase, scGPT learns both cell and gene embeddings, which capture biological variations and interactions within the data.

For downstream applications, such as cell type annotation, multi-omic integration, or perturbation response prediction, the pretrained model is fine-tuned on specific datasets using supervised objectives. For cell type annotation, for example, a neural network classifier is added on top of the pretrained cell embeddings, and the whole model is trained using cross-entropy loss. The fine-tuned model is then used to predict cell types on new, unseen datasets, demonstrating the model's capability for transfer learning.

The scGPT model also enables multi-batch and multi-omic integration by learning unified cell representations that can recover masked gene expression patterns across different modalities. For gene regulatory network (GRN) inference, the attention weights and learned gene embeddings can be utilized to identify interactions and regulatory relationships between genes, making scGPT a versatile tool for various single-cell analysis tasks.

The mathematical formulation for the masked language modeling objective used during pretraining is as follows:

\[
\mathcal{L}_{\text{MLM}} = - \sum_{i \in \mathcal{M}} \log P(x_i \mid \hat{x}_{\mathcal{M} \backslash \{i\}}, \theta)
\]

where \(\mathcal{M}\) is the set of masked gene indices, \(x_i\) is the masked gene expression value, and \(\hat{x}_{\mathcal{M} \backslash \{i\}}\) denotes the remaining observed gene expression values. The model parameters \(\theta\) are optimized to maximize the likelihood of recovering the masked values.

\subsection{CellPLM Model}

CellPLM (Cell Pre-trained Language Model) \citep{chen2021single} is a novel framework designed to address the challenges in single-cell transcriptomic data analysis by leveraging concepts from natural language processing (NLP). It models cells as tokens and tissues as sentences, incorporating both gene expression data and spatially-resolved transcriptomic (SRT) data to capture complex cell-cell relationships.

The architecture of CellPLM consists of four primary components: a \textbf{gene expression embedder}, an \textbf{encoder}, a \textbf{Gaussian mixture latent space}, and a \textbf{decoder}.

The gene expression embedder projects input gene expressions into a low-dimensional feature space. Each gene type $j$ is assigned a learnable embedding vector $h_j \in \mathbb{R}^d$, where $d$ is the hidden dimension. For each cell $i$, the gene expression embedding matrix $E_i$ is computed as:
\begin{equation}
    E_i = \sum_{j=1}^{k} X_{i,j} h_j,
\end{equation}
where $X \in \mathbb{R}^{N \times k}$ represents the cell-by-gene expression matrix, $N$ is the number of cells, $k$ is the number of gene types, and $X_{i,j}$ denotes the expression level of gene $j$ in cell $i$.

CellPLM uses a transformer encoder to capture intercellular relationships by treating cells as tokens. The encoder operates on a set of cell embeddings and applies multi-head self-attention mechanisms. Given $L$ stacked transformer layers, the representation of cells at layer $l$ is:
\begin{equation}
    H^{(l)} = \text{TransformerLayer}^{(l)}(H^{(l-1)}),
\end{equation}
where $H^{(l-1)}$ is the cell representation from the previous layer. The input representation $H^{(0)}$ is the sum of gene expression embeddings $E$ and positional encodings $P$.

The latent space in CellPLM employs a Gaussian mixture model to better capture biological structures and overcome batch effects. The generative process for each cell $i$ is defined as:
\begin{align}
    p(y_i; \pi) &= \text{Multinomial}(\pi), \\
    p(z_i \mid y_i) &= \prod_{l=1}^{L} \mathcal{N}(\mu_{y_i,l}, \text{diag}(\sigma^2_{y_i,l})), \\
    p_{\theta_{\text{dec}}}(x_i \mid z_i) &= \mathcal{N}(\mu_{z_i}, \sigma^2 I),
\end{align}
where $y_i$ is the latent cluster variable, $\pi$ is its prior, $\mu_{y_i,l}$ and $\sigma^2_{y_i,l}$ are the mean and variance of the $l$-th Gaussian component, and $z_i$ is the latent variable for cell $i$.

The decoder reconstructs masked features and removes batch effects by using a series of feed-forward layers. It incorporates a batch embedding $b$ for each cell, formulated as:
\begin{equation}
    h^{(0)} = z + b, \quad h^{(l)} = \text{FFLayer}^{(l)}(h^{(l-1)}),
\end{equation}
where $h^{(l)}$ is the hidden vector at layer $l$.

\textbf{Pre-training} is performed on a masked language modeling task, where certain gene expressions are masked, and the model is trained to reconstruct the original values. The objective function is defined as:
\begin{equation}
    L_{\text{CellPLM}} = L_{\text{recon}} + L_{\text{cond}} + L_{Y},
\end{equation}
where $L_{\text{recon}}$ is the reconstruction loss, $L_{\text{cond}}$ ensures the consistency between the posterior and conditional prior, and $L_{Y}$ aligns the latent cluster distribution with the prior. The reconstruction loss is computed as:
\begin{equation}
    L_{\text{MSE}} = \left\| M \odot \left( H^{(L)} - (1-M) \odot X \right) \right\|_F^2,
\end{equation}
where $M$ is the mask indicator matrix, $H^{(L)}$ is the output from the decoder, and $\odot$ denotes element-wise multiplication.

\textbf{Fine-tuning} uses the pre-trained parameters as initialization and adapts them to specific downstream tasks, such as cell type annotation and spatial transcriptomic imputation, using task-specific loss functions.

For SRT data, CellPLM incorporates additional positional information through 2D coordinates of cells, which is encoded into a positional encoding matrix $P \in \mathbb{R}^{N \times d}$. This enables the model to integrate spatial context into the cell representations, facilitating the understanding of cell-cell interactions.

\subsection{Geneformer Model}

Geneformer \citep{chen2023geneformer} is a context-aware, attention-based deep learning model designed for predictions in network biology using transfer learning. The model was pretrained on a corpus of 29.9 million single-cell transcriptomes, known as \textit{Genecorpus-30M}, from a broad range of tissues using a self-supervised masked learning objective. During pretraining, Geneformer learned to encode network hierarchies and gene interactions by predicting masked genes within the context of unmasked genes for each cell state. This approach enabled the model to capture fundamental network dynamics in a completely unsupervised manner, which can then be transferred to diverse downstream tasks that may have limited data.

The architecture of Geneformer consists of six transformer encoder layers. Each encoder layer contains a self-attention layer and a feed-forward neural network layer. The model uses full dense self-attention over an input size of 2,048 genes, representing 93\% of rank value encodings in the Genecorpus-30M. Each gene is encoded into a 256-dimensional space, and Geneformer uses a rank value encoding where genes are ranked by their expression levels, normalized across the entire dataset. This rank-based encoding provides a non-parametric representation of the transcriptome of each single cell, deprioritizing housekeeping genes while prioritizing genes that distinguish specific cell states, such as transcription factors.

The self-supervised pretraining was accomplished by masking 15\% of the genes within each transcriptome and training the model to predict the masked genes. This objective is similar to the masked language modeling objective used in natural language processing. The pretraining allows the model to learn which genes are most relevant within a cell's context, thus establishing an understanding of network dynamics that is robust to technical noise and batch effects.

The Geneformer model is implemented with self-attention mechanisms, where each attention head learns which genes to focus on in different contexts. During training, the attention weights are optimized iteratively to generate gene embeddings that best inform the correct answer for the learning objective. Specifically, Geneformer uses multiple attention heads per layer, each capturing distinct aspects of the gene regulatory networks, allowing the model to pay attention to diverse features simultaneously.

Mathematically, given an input transcriptome matrix \( X \in \mathbb{R}^{N \times d} \), where \( N \) is the number of genes and \( d \) is the embedding dimension, Geneformer applies self-attention mechanisms as follows:

\[
\text{Attention}(Q, K, V) = \text{softmax} \left( \frac{QK^T}{\sqrt{d_k}} \right) V
\]

where \( Q \), \( K \), and \( V \) are the query, key, and value matrices derived from the input matrix \( X \). The attention scores are computed by taking the dot product of \( Q \) and \( K \), scaled by \( \sqrt{d_k} \) (the square root of the key dimension) and normalized with a softmax function.

The embeddings are then passed through multiple transformer encoder layers, each performing layer normalization, multi-head attention, and position-wise feed-forward neural networks:

\[
\text{FFN}(x) = \text{max}(0, xW_1 + b_1)W_2 + b_2
\]

The final output of the pretraining is a set of contextualized gene and cell embeddings that can be used for downstream analyses, such as predicting gene dosage sensitivity or chromatin dynamics.

\subsection{Architecture Implementation of scVI}

Single-cell Variational Inference (scVI) \citep{scvilopez2018deep} is a probabilistic model designed for analyzing single-cell RNA sequencing (scRNA-seq) data, addressing challenges like technical noise, dropout events, and batch effects. The model is based on a hierarchical Bayesian framework, leveraging deep neural networks for representation learning and scalable variational inference.

The input to scVI is an \( N \times G \) count matrix \( \mathbf{x} \) that records the number of transcripts measured for \( G \) genes across \( N \) cells. To account for technical variation, such as differences in capture efficiency or sequencing depth, scVI introduces two types of latent variables:

\begin{enumerate}
    \item \textbf{Biological Latent Variable \( \mathbf{z}_n \):} Represents the low-dimensional latent space capturing the biological variability between cells. This variable is drawn from a standard multivariate normal distribution:

\[
\mathbf{z}_n \sim \mathcal{N}(\mathbf{0}, \mathbf{I}).
\]

    \item \textbf{Scaling Factor \( \ell_n \):} Models cell-specific variation due to sequencing depth or capture efficiency. It is drawn from a log-normal distribution:
\[
\ell_n \sim \text{LogNormal}(\mu_\ell, \sigma_\ell^2).
\]

\end{enumerate}

The generative process of scVI can be summarized as follows:

\begin{itemize}
    \item Sample \( \mathbf{z}_n \) and \( \ell_n \) for each cell \( n \).
    \item Compute a batch-corrected representation \( \boldsymbol{\rho}_{gn} \) for each gene \( g \) and cell \( n \), using neural networks \( f_w \) and \( f_h \):
    \[
    \boldsymbol{\rho}_{gn} = f_w(\mathbf{z}_n, s_n),
    \]
    where \( s_n \) is the batch annotation for cell \( n \).
    \item Draw \( w_{ng} \) from a gamma distribution:
    \[
    w_{ng} \sim \Gamma(\boldsymbol{\rho}_{gn}, \theta_g),
    \]
    where \( \theta_g \) is a gene-specific inverse dispersion parameter.
    \item Given \( \ell_n \) and \( w_{ng} \), sample \( y_{ng} \) from a Poisson distribution:
    \[
    y_{ng} \sim \text{Poisson}(\ell_n w_{ng}).
    \]
    \item Include zero-inflation to account for dropout events by introducing a Bernoulli variable \( h_{ng} \) for each gene \( g \) and cell \( n \):
    \[
    h_{ng} \sim \text{Bernoulli}(f_h(\mathbf{z}_n, s_n)).
    \]
    \item The observed expression count \( x_{ng} \) is defined as:
    \[
    x_{ng} = 
    \begin{cases}
    y_{ng} & \text{if } h_{ng} = 0, \\
    0 & \text{if } h_{ng} = 1.
    \end{cases}
    \]
\end{itemize}

In essence, the model assumes that the expression values \( x_{ng} \) are generated through a zero-inflated negative binomial (ZINB) distribution:
\[
x_{ng} \sim \text{ZINB}(\ell_n \boldsymbol{\rho}_{gn}, \theta_g, f_h(\mathbf{z}_n, s_n)).
\]
Due to the non-conjugacy and complex nature of this distribution, exact Bayesian inference is not feasible. Therefore, scVI uses a variational approximation \( q(\mathbf{z}_n, \log \ell_n | \mathbf{x}_n, s_n) \) with a Gaussian form. The variational parameters are optimized using stochastic gradient descent.

\section{Hierarchy of metrics for Transcriptomics Benchmarking}


\subsection{Formulation of the iLISI Score}

The Integration Local Inverse Simpson’s Index (iLISI) is a metric used to quantify the extent to which samples from different batches are well integrated within the latent space of a model. In biological experiments, data is often collected in multiple \textit{batches}, where each batch consists of samples processed under slightly different conditions or at different times. These batch differences can introduce \textit{batch effects}, which are unwanted variations that can obscure true biological signals. The iLISI score measures how well a model reduces these batch effects by assessing the mixing of samples from different batches in the neighborhood of each cell or sample.

The iLISI score is computed as follows:

\textbf{1. Calculate Neighborhood Probabilities:} For each sample \( i \), identify its set of \( k \)-nearest neighbors, denoted as \( \mathcal{N}_i \). The distance between samples \( i \) and \( j \) in the latent space is represented by \( d_{ij} \). Using this distance, the conditional probability \( p_{ij} \) that sample \( i \) selects sample \( j \) as a neighbor is defined as:
    \[
    p_{ij} = \frac{\exp(-\beta_i d_{ij})}{\sum_{l \in \mathcal{N}_i} \exp(-\beta_i d_{il})},
    \]
    
where \( d_{ij} \) is the distance between sample \( i \) and sample \( j \), and \( \beta_i \) is the scaling parameter that ensures the entropy \( H(P_i) \) of the distribution \( P_i = \{ p_{ij} \} \) is equal to \( \log(k) \). This scaling ensures that the neighborhood size matches the target number of neighbors \( k \).

\textbf{2. Compute Batch Probability for Each Sample:} Let \( C \) be the set of all batch labels, and \( l_j \) denote the batch label of neighbor \( j \). For each sample \( i \), calculate the sum of probabilities for neighbors that belong to batch \( c \in C \):
    \[
    p_{ic} = \sum_{\substack{j \in \mathcal{N}_i \\ l_j = c}} p_{ij}.
    \]
    This value \( p_{ic} \) represents the probability that a neighbor of sample \( i \) is from batch \( c \).

\textbf{3. Calculate the Inverse Simpson’s Index:} For each sample \( i \), the Inverse Simpson’s Index \( \text{ISI}_i \) is computed as:
    \[
    \text{ISI}_i = \left( \sum_{c \in C} p_{ic}^2 \right)^{-1}.
    \]
    The Inverse Simpson’s Index measures the diversity of batches in the neighborhood of each sample. A high value indicates that the neighbors are well-distributed across different batches, suggesting effective mixing and reduced batch effects.

\textbf{4. Compute the Final iLISI Score:} The overall iLISI score is the average of \( \text{ISI}_i \) over all \( n \) samples:
    \[
    \text{iLISI} = \frac{1}{n} \sum_{i=1}^{n} \text{ISI}_i = \frac{1}{n} \sum_{i=1}^{n} \left( \sum_{c \in C} p_{ic}^2 \right)^{-1}.
    \]

The iLISI score provides an indication of how well the batches are mixed in the latent space. An ideal iLISI score would be close to the number of unique batches in the dataset, before normalization. This would indicate that, on average, the neighbors of each sample are evenly distributed across all batches, implying that the model has effectively minimized batch effects. Conversely, a low iLISI score would suggest poor mixing, indicating that the model has not successfully integrated samples from different batches.

\subsection{Formulation of the Perturbation Consistency Score}

The \textit{Perturbation Consistency} score is designed to evaluate how consistently a model represents the effect of a perturbation across different samples and batches. This metric is crucial for ensuring that the model’s representations of perturbations are robust, reliable, and not influenced by random noise or technical artifacts. The Perturbation Consistency score quantifies the similarity of the model’s embeddings for samples subjected to the same perturbation, thereby providing a measure of how well the perturbation’s effect is captured and maintained throughout the representation space.

The computation of the Perturbation Consistency score involves the following steps:

\textbf{1. Calculate Pairwise Cosine Similarities for Each Perturbation:} For each perturbation \( g \), represented as a set of samples, we compute the cosine similarity between the embeddings of every pair of samples. Let \( \mathbf{x}_{g,i} \) be the embedding vector for the \( i \)-th sample of perturbation \( g \). The cosine similarity between two sample embeddings \( \mathbf{x}_{g,i} \) and \( \mathbf{x}_{g,j} \) is given by:
    \[
    \text{cos}(\mathbf{x}_{g,i}, \mathbf{x}_{g,j}) = \frac{\langle \mathbf{x}_{g,i}, \mathbf{x}_{g,j} \rangle}{\|\mathbf{x}_{g,i}\| \, \|\mathbf{x}_{g,j}\|},
    \]
    where \( \langle \mathbf{x}_{g,i}, \mathbf{x}_{g,j} \rangle \) denotes the dot product between \( \mathbf{x}_{g,i} \) and \( \mathbf{x}_{g,j} \), and \( \|\mathbf{x}_{g,i}\| \) is the Euclidean norm of \( \mathbf{x}_{g,i} \).

\textbf{2. Compute the Average Cosine Similarity:} For each perturbation \( g \), with \( n_g \) samples, the average cosine similarity \( \text{avgsim}_g \) is computed by averaging over all pairs of embeddings:
    \[
    \text{avgsim}_g = \frac{1}{n_g^2} \sum_{i=1}^{n_g} \sum_{j=1}^{n_g} \text{cos}(\mathbf{x}_{g,i}, \mathbf{x}_{g,j}).
    \]
    This average similarity score provides a measure of how close the representations of samples under the same perturbation are, indicating the degree of consistency.

\textbf{3. Comparison to Null Distribution:} To determine if the observed consistency score \( \text{avgsim}_g \) is significant, it is compared to a null distribution generated from unexpressed or inactive genes. These genes are selected based on consistently low expression levels across all samples, and their embeddings are used to compute a null average similarity \( \text{avgsim}_{g_k'} \) for each unexpressed gene \( g_k' \), where \( k = 1, \dots, K \).

\textbf{4. Permutation Test:} For each perturbation \( g \), the consistency $p$-value is calculated using a permutation test. The $p$-value is defined as the fraction of unexpressed genes with average similarity scores less than or equal to \( \text{avgsim}_g \):
    \[
    p_g = \frac{\max\left\{ \#\left( \text{avgsim}_{g_k'} \leq \text{avgsim}_g \right), 1 \right\}}{K}.
    \]
    This step ensures that the significance of the observed similarity is evaluated against what would be expected by chance.

\textbf{5. Final Perturbation Consistency Score:} Gene perturbations with a consistency $p$-value less than 0.05 are considered significant, indicating that the perturbation's effect is consistently represented across different samples and batches. The overall Perturbation Consistency score is reported as the fraction of significant gene perturbations out of the total number of perturbations evaluated. It is formally defined as:
    \[
    \text{Perturbation Consistency} = \frac{\#(\text{Significant Perturbations})}{\#(\text{Total Perturbations})}.
    \]

A high Perturbation Consistency score indicates that the model reliably captures the effect of each perturbation across all samples, suggesting that the model’s representations are robust to noise and batch effects. In contrast, a low score suggests that the model may fail to capture the perturbation effect consistently, potentially due to noise, batch-specific artifacts, or poor representation learning. Thus, this metric is essential for validating the stability and reliability of a model's performance in perturbation analysis.

\subsection{Formulation of the Structural Integrity Score}

The \textit{Structural Integrity} score is a metric designed to evaluate how well a model preserves the relationship between control and perturbation conditions within each biological batch in terms of gene activity. This score is particularly useful for analyzing gene expression changes under different conditions by comparing the predicted and actual gene expression profiles. The structural integrity score is computed based on the Frobenius norm of the difference between the centered predicted and centered actual gene expression matrices for each batch, providing a measure of how accurately the model reconstructs gene expression while accounting for batch-specific variability.

The computation of the Structural Integrity score involves the following steps:

\textbf{1. Centering the Gene Expression Profiles:} For each batch \( b \), the (log-transformed) gene expression profiles are centered by subtracting the corresponding control condition within that batch. This operation is performed separately for both the predicted gene expression matrix \( Y_{\text{pred}}^{(b)} \) and the actual gene expression matrix \( Y_{\text{actual}}^{(b)} \). The centered matrices are denoted as \( \tilde{Y}_{\text{pred}}^{(b)} \) and \( \tilde{Y}_{\text{actual}}^{(b)} \), respectively:
    \[
    \tilde{Y}_{\text{pred}}^{(b)} = Y_{\text{pred}}^{(b)} - Y_{\text{pred, control}}^{(b)}, \quad \tilde{Y}_{\text{actual}}^{(b)} = Y_{\text{actual}}^{(b)} - Y_{\text{actual, control}}^{(b)},
    \]
    where \( Y_{\text{pred, control}}^{(b)} \) and \( Y_{\text{actual, control}}^{(b)} \) represent the control condition matrices for batch \( b \).

\textbf{2. Compute Structural Distance:} The structural distance quantifies the deviation between the centered predicted and actual gene expression matrices for each batch \( b \). It is defined as the Frobenius norm of the difference between these two matrices, averaged over all batches:
    \[
    \text{Structural Distance} = \frac{1}{B} \sum_{b=1}^{B} \frac{1}{n_b} \left\| \tilde{Y}_{\text{pred}}^{(b)} - \tilde{Y}_{\text{actual}}^{(b)} \right\|_F,
    \]
    where \( B \) is the total number of batches, \( n_b \) is the number of samples in batch \( b \), and \( \|\cdot\|_F \) denotes the Frobenius norm, which is computed as:
    \[
    \left\| A \right\|_F = \sqrt{\sum_{i=1}^{m} \sum_{j=1}^{n} a_{ij}^2},
    \]
    for any matrix \( A \in \mathbb{R}^{m \times n} \).

\textbf{3. Derivation of the Maximum Structural Distance:} The Structural Distance measures the average discrepancy between the centered predicted and actual gene expression profiles across all biological batches. To derive the upper bound, we consider the maximum possible difference between these profiles. After centering the gene expression profiles by subtracting the corresponding control within each batch, the maximum possible value for any element in the adjusted matrices \( \tilde{Y}_{\text{pred}}^{(b)} \) and \( \tilde{Y}_{\text{actual}}^{(b)} \) depends on the data but can be bounded. Assuming the gene expression values are normalized and bounded within a range \([0, M]\), the centered gene expression values are bounded within a range \([-M, M]\), thus the maximum difference per element between the centered predicted and actual matrices is \( 2M \).

For each batch \( b \), the maximum possible Structural Distance is given by:

\[
\text{Max Structural Distance}_b = \left\| \tilde{Y}_{\text{pred}}^{(b)} - \tilde{Y}_{\text{actual}}^{(b)} \right\|_F^{\max} = 2M \sqrt{n_b \times g},
\]

where \( n_b \) is the number of samples in batch \( b \), \( g \) is the number of genes and \( \|\cdot\|_F \) denotes the Frobenius norm.

The overall maximum Structural Distance across all batches is then computed as:

\[
\text{Structural Distance}_{\text{max}} = \frac{1}{B} \sum_{b=1}^{B} 2M \sqrt{n_b \times g}.
\]

This theoretical maximum serves as a normalization factor to evaluate the structural distance relative to an upper bound. This measure can be approximated to the distance of centered actual gene expression profiles to their negative value, which can be defined as follow : 

\begin{equation}
    \text{Structural Distance}_{max} \approx \frac{2}{B} \sum_{b=1}^{B} \frac{1}{n_b} \left\| \tilde{Y}_{\text{actual}}^{(b)} \right\|_F,
\end{equation}

\textbf{4. Compute the Structural Integrity Score:} The Structural Integrity score is calculated as:
    \[
    \text{Structural Integrity} = 1 - \frac{\text{Structural Distance}}{\text{Structural Distance}_{\text{max}}}.
    \]
    This formula provides a normalized measure of how well the model preserves the structural relationships in gene expression data. A Structural Integrity score close to 1 indicates high preservation of the control-perturbation relationship, while a score closer to 0 indicates poor preservation.

\section{Experimental Setup}

\subsection{Embedding Post-Processing}

\subsubsection{Centering}
Centering is the process of adjusting the data such that the mean of each feature (or dimension) is zero. This operation is performed by subtracting the mean value of each feature from the data. For a given feature matrix \( X \in \mathbb{R}^{n \times m} \), where \( n \) is the number of samples and \( m \) is the number of features, the centered matrix \( \tilde{X} \) is obtained as:
\[
\tilde{X}_{ij} = X_{ij} - \frac{1}{n} \sum_{k=1}^{n} X_{kj}, \quad \forall i = 1, \ldots, n, \quad \forall j = 1, \ldots, m.
\]
This transformation removes the mean from each feature, thus translating the data so that it is centered around zero. In the context of batch-corrected biological data, centering is often applied to the final embeddings to remove the negative control embeddings, allowing focus on perturbation effects.

\subsubsection{Center Scaling}
Center scaling further extends centering by normalizing the variance of each feature. After centering, the data is scaled such that each feature has unit variance. For a centered feature matrix \( \tilde{X} \), the scaled matrix \( \hat{X} \) is obtained as:
\[
\hat{X}_{ij} = \frac{\tilde{X}_{ij}}{\sigma_j}, \quad \sigma_j = \sqrt{\frac{1}{n} \sum_{k=1}^{n} \tilde{X}_{kj}^2}, \quad \forall i = 1, \ldots, n, \quad \forall j = 1, \ldots, m,
\]
where \( \sigma_j \) is the standard deviation of the \( j \)-th feature. By performing both centering and scaling, the features become comparable, which is essential when using techniques such as Principal Component Analysis (PCA) that are sensitive to the scale of the data.

\subsubsection{Typical Variation Normalization (TVN)}
Typical Variation Normalization (TVN) is a normalization technique designed to enhance the representation of biological data by reducing batch effects and highlighting subtle phenotypic differences. TVN is particularly useful in high-content imaging screens and other applications where there is substantial variability between experimental batches.

TVN operates by first computing the principal components of control samples (negative control conditions) to identify the axes of typical variation in the data. The principal component analysis (PCA) is performed on the centered control embeddings \( \tilde{X}_{\text{control}} \), resulting in a set of principal components \( \{ \mathbf{v}_1, \ldots, \mathbf{v}_m \} \). Each component represents a direction in the data space that explains a certain amount of variance. The principal components are then used to normalize the embeddings as follows:

\textbf{1. Centering and Scaling of Negative Controls:}
    The negative control embeddings \( \tilde{X}_{\text{control}} \) are centered and scaled using their respective means and standard deviations, such that:
    \[
    \hat{X}_{\text{control}} = \frac{\tilde{X}_{\text{control}} - \mu_{\text{control}}}{\sigma_{\text{control}}},
    \]
    where \( \mu_{\text{control}} \) and \( \sigma_{\text{control}} \) are the mean and standard deviation of the negative control embeddings, respectively.

\textbf{2. Principal Component Analysis (PCA):}
    PCA is applied to \( \hat{X}_{\text{control}} \) to compute the principal components. Let \( W \in \mathbb{R}^{m \times m} \) be the matrix whose columns are the principal component vectors \( \mathbf{v}_j \).

\textbf{3. TVN Transformation:}
    The transformation matrix \( T \) is derived from the principal components such that the variance along each axis is normalized:
    \[
    T = W \cdot D^{-1/2} \cdot W^\top,
    \]
    where \( D \) is a diagonal matrix containing the eigenvalues associated with the principal components.

\textbf{4. Application to All Embeddings:}
    The TVN transformation is applied to both control and experimental condition embeddings \( X_{\text{all}} \):
    \[
    X_{\text{TVN}} = T \cdot X_{\text{all}}.
    \]
    This step reduces the impact of nuisance variation in the embeddings, ensuring that the embeddings are desensitized to batch-specific differences while amplifying the axes of variation that capture subtle or rare biological phenotypes.

\subsection{Data Format and Splits}

We utilize three primary open-source datasets in this study: the single-cell gene knockout dataset \citep{Replogle2022-lx}, the bulk L1000 CRISPR assay dataset \citep{Subramanian2017-rx}, and the single-cell gene overexpression dataset \citep{Joung2023-nr}. Each dataset captures distinct aspects of gene perturbation and expression, providing a comprehensive benchmark for evaluating the models under different experimental conditions and perturbation types.

The single-cell gene knockout dataset \citep{Replogle2022-lx} contains 1.98 million samples derived from a single cell line (K562), covering 9,866 genetic knockouts across 267 different biological batches/experiments. The dataset includes 75,328 control samples and measures the expression of 8,248 genes. In contrast, the bulk L1000 CRISPR assay dataset \citep{Subramanian2017-rx} consists of 443,365 samples from 31 different cell lines, including 5,157 genetic knockouts and 87,565 control samples across 1,188 batches. This dataset directly measures around 1,000 genes and infers the expression of an additional 11,322 genes. Finally, the single-cell gene overexpression dataset \citep{Joung2023-nr}, designed for out-of-distribution training, includes 1.38 million samples from a single cell line (H1 hESC), covering 1,762 genetic overexpression perturbations applied across two batches/experiments and measuring 37,528 genes. It also contains 173,211 control samples. This dataset is primarily used for training models on gene overexpression before evaluating the trained model on gene knockout in different cell lines.

To maintain consistency, all datasets are preprocessed to use the same metadata and structure, ensuring compatibility during model training and evaluation. Only samples with total raw counts above 1,000 are retained for analysis. For each dataset, we split the data into training and test sets based on distinct biological batches. This approach ensures that the test set contains batches not seen during training, minimizing potential information leakage and providing a more rigorous evaluation of the model’s ability to generalize to new conditions. For linear probing and kNN evaluation, we employ a 70\% train/test split, ensuring that the test set contains distinct biological batches while retaining the same perturbations as the training set. In contrast, the reconstruction test set is designed to evaluate the model's capacity to generalize to new perturbations. It contains distinct perturbations and batches that were not present in the training set. The complete datasets are used for perturbation consistency and zero-shot known relationship recall tasks. 

For the L1000 assay dataset, the training set comprises 234,571 samples from 831 distinct biological batches, including 61,839 control samples and 3,603 unique perturbations. The test set used for linear probing and k-Nearest Neighbors (kNN) evaluation consists of 100,153 samples from 356 distinct batches, with 25,691 control samples and the same 3,603 perturbations as in the training set. A separate reconstruction test set is used for evaluating representation learning. This reconstruction set includes 57,956 samples from 354 distinct batches, with 25,572 control samples and 1,549 unique perturbations not present in the training set. For the Replogle single-cell gene knockout dataset \citep{Replogle2022-lx}, the training set includes 992,027 samples from 186 biological batches, with 52,900 control samples and 6,905 unique perturbations. The linear probing and kNN test set consists of 424,942 samples from 81 distinct batches, with 22,428 control samples and the same 6,905 perturbations as the training set. Similar to the L1000 data, a separate reconstruction test set is utilized, containing 194,178 samples from 81 distinct batches, with 22,428 control samples and 2,960 unique perturbations that are not present in the training set. The single-cell gene overexpression dataset \citep{Joung2023-nr} is primarily used for training models on gene overexpression perturbations before evaluating them on gene knockout tasks. This dataset is not split further for evaluation purposes as its primary role is to serve as an out-of-distribution training set. The generalization of models trained on this dataset is assessed using the gene knockout datasets.

\subsection{PCA, scVI and finetuning scGPT}

For training the scVI model, we configure it with a latent space size of 256 and set the hidden space dimension to \( n_{\text{hidden}} \) with 2 layers. The number of highly variable genes selected differs based on the dataset: 15,000 genes for the Joung overexpression dataset, 8,000 genes for the Replogle dataset, and 12,000 genes for the L1000 assay. The dispersion parameter is set to "gene", and the latent distribution follows a normal distribution. We test various gene likelihood models, including Zero-Inflated Negative Binomial (ZINB), Negative Binomial (NB), and Poisson distributions. To ensure consistent comparison, we normalize the data to a library size of 10,000 counts per sample and apply log-transformation prior to training scVI. The normalization and transformation procedures help mitigate the effect of varying library sizes and improve model stability during training.

For data analysis using Principal Component Analysis (PCA), we apply the PCA transformation on the gene expression profiles prior to normalization and log transformation. This ensures that the PCA captures the intrinsic variability in the raw expression data without additional preprocessing artifacts.

To fine-tune the scGPT model, we configure the model with an embedding dimension of 512, a hidden layer size of 512, and 12 Transformer layers with 8 attention heads. The input sequence is padded to a maximum length of 1,536, and special tokens such as "<pad>", "<cls>", and "<eoc>" are included in the vocabulary. The training uses a learning rate of \( 10^{-5} \) with a batch size of 32 and is carried out for 15 epochs. The scGPT model is trained using masked language modeling (MLM), with no additional objectives such as cell-type classification or contrastive cell embedding. The optimizer used is Adam, and a learning rate scheduler is applied at an interval of 1 epoch, with a decay factor of 0.9. Gradient scaling is enabled using mixed precision training to accelerate computations and reduce memory usage. 5000 most variable genes were used for training.

\subsection{Evaluation Parameters}

For training on the L1000 and Replogle datasets, distinct configurations are used for kNN, linear evaluation, and reconstruction tasks. During kNN and linear probing, the training data is fed with a batch size of 2048, and optimization is performed using AdamW with a learning rate of 0.001 and a weight decay of \(1 \times 10^{-6}\). The learning rate follows a warmup cosine decay schedule, starting at \(3 \times 10^{-5}\) for the initial 10 epochs and decaying to zero over 250 epochs. The model is set to evaluate its performance using a linear classifier, and leverages 251 neighbors for kNN evaluation. For both datasets, We perform training with five distinct but fixed seeds to ensure reproducibility, and each training is conducted using mixed precision on a single H100 GPU. We take the average and standard deviation across the 5 training runs with distinct seeds.

Reconstruction tasks use a smaller batch size of 512 samples and a lower learning rate of 0.0001. The model undergoes training for a maximum of 30 epochs, utilizing the same optimizer and warmup cosine learning rate schedule as in linear probing. We use two stacked linear layers to decode the latent space into into log transformed gene expression profiles, and MSE loss for the reconstruction.

\section{Detailed Results}


\begin{table}[h]
    \centering
    \caption{\cite{Replogle2022-lx} dataset full results}
    \label{tab:model_performance}
    \begin{tabular}{llccc}
    \toprule
    \textbf{Model Type} & \textbf{Processing Method} & \textbf{iLISI} & \textbf{Top1 Linear Acc} & \textbf{Top5 Linear Acc} \\
    \midrule
    \multirow{4}{*}{Geneformer} 
    & Center Scaling & 21.361 ± 0.051 & 0.021 ± 0.001 & 0.101 ± 0.003 \\
    & Centering & 22.069 ± 0.028 & 0.112 ± 0.007 & 0.490 ± 0.007 \\
    & Raw Embeds & 22.273 ± 0.017 & 0.155 ± 0.001 & 0.536 ± 0.002 \\
    & TVN & 6.845 ± 1.114 & 0.015 ± 0.001 & 0.083 ± 0.002 \\
    \midrule
    \multirow{4}{*}{PCA} 
    & Center Scaling & 20.909 ± 0.497 & 3.269 ± 0.004 & 5.765 ± 0.007 \\
    & Centering & 16.789 ± 0.043 & 3.094 ± 0.005 & 5.480 ± 0.008 \\
    & Raw Embeds & 15.016 ± 0.066 & 3.221 ± 0.008 & 5.668 ± 0.004 \\
    & TVN & 16.637 ± 0.477 & 2.922 ± 0.006 & 5.219 ± 0.004 \\
    \midrule
    \multirow{4}{*}{Transfer scVI} 
    & Center Scaling & 20.041 ± 0.084 & 0.371 ± 0.002 & 0.895 ± 0.003 \\
    & Centering & 12.436 ± 0.189 & 0.384 ± 0.001 & 0.930 ± 0.003 \\
    & Raw Embeds & 14.887 ± 0.158 & 0.389 ± 0.002 & 0.973 ± 0.005 \\
    & TVN & 16.401 ± 1.096 & 0.300 ± 0.002 & 0.759 ± 0.001 \\
    \midrule
    \multirow{4}{*}{UCE} 
    & Center Scaling & 21.187 ± 0.033 & 0.205 ± 0.002 & 0.572 ± 0.002 \\
    & Centering & 21.185 ± 0.037 & 0.262 ± 0.002 & 0.772 ± 0.002 \\
    & Raw Embeds & 17.495 ± 0.015 & 0.224 ± 0.001 & 0.725 ± 0.004 \\
    & TVN & 11.409 ± 1.073 & 0.065 ± 0.001 & 0.206 ± 0.001 \\
    \midrule
    \multirow{4}{*}{cellPLM} 
    & Center Scaling & 11.879 ± 0.036 & 0.274 ± 0.002 & 0.837 ± 0.004 \\
    & Centering & 19.091 ± 0.037 & 0.247 ± 0.004 & 0.803 ± 0.005 \\
    & Raw Embeds & 17.952 ± 0.017 & 0.234 ± 0.002 & 0.770 ± 0.005 \\
    & TVN & 14.883 ± 0.107 & 0.313 ± 0.003 & 0.873 ± 0.003 \\
    \midrule
    \multirow{4}{*}{scGPT} 
    & Center Scaling & 21.407 ± 0.024 & 0.438 ± 0.004 & 1.113 ± 0.005 \\
    & Centering & 20.561 ± 0.033 & 0.269 ± 0.002 & 0.809 ± 0.002 \\
    & Raw Embeds & 15.935 ± 0.027 & 0.231 ± 0.001 & 0.750 ± 0.003 \\
    & TVN & 8.216 ± 1.147 & 0.191 ± 0.002 & 0.506 ± 0.002 \\
    \midrule
    \multirow{4}{*}{scGPT finetuned} 
    & Center Scaling & 20.665 ± 0.016 & 0.037 ± 0.000 & 0.167 ± 0.002 \\
    & Centering & 20.756 ± 0.021 & 0.118 ± 0.004 & 0.480 ± 0.007 \\
    & Raw Embeds & 21.707 ± 0.017 & 0.155 ± 0.001 & 0.533 ± 0.001 \\
    & TVN & 10.038 ± 1.473 & 0.018 ± 0.000 & 0.104 ± 0.001 \\
    \midrule
    \multirow{4}{*}{scVI} 
    & Center Scaling & 19.869 ± 0.146 & 1.589 ± 0.005 & 3.190 ± 0.003 \\
    & Centering & 14.842 ± 0.141 & 1.607 ± 0.004 & 3.218 ± 0.002 \\
    & Raw Embeds & 14.896 ± 0.099 & 1.566 ± 0.005 & 3.128 ± 0.002 \\
    & TVN & 15.088 ± 1.598 & 1.430 ± 0.005 & 2.896 ± 0.004 \\
    \bottomrule
    \end{tabular}
\end{table}


\begin{table}[ht]
    \centering
    \caption{\cite{Replogle2022-lx} dataset full results}
    \label{tab:model_performance}
    \begin{tabular}{llccc}
    \toprule
    \textbf{Model Type} & \textbf{Processing Method} & \textbf{Pert Consistency} & \textbf{Top1 KNN Acc} & \textbf{Top5 KNN Acc} \\
    \midrule
    \multirow{4}{*}{Geneformer} 
    & Center Scaling & 6.000 ± 0.000 & 5.280 ± 0.007 & 5.384 ± 0.006 \\
    & Centering & 5.200 ± 0.000 & 5.280 ± 0.006 & 5.381 ± 0.008 \\
    & Raw Embeds & 5.300 ± 0.000 & 5.280 ± 0.007 & 5.384 ± 0.006 \\
    & TVN & 6.300 ± 0.000 & 5.281 ± 0.007 & 5.365 ± 0.007 \\
    \midrule
    \multirow{4}{*}{PCA} 
    & Center Scaling & 11.900 ± 0.000 & 5.348 ± 0.017 & 5.865 ± 0.055 \\
    & Centering & 9.400 ± 0.000 & 5.375 ± 0.008 & 6.136 ± 0.008 \\
    & Raw Embeds & 9.200 ± 0.000 & 5.382 ± 0.006 & 6.111 ± 0.010 \\
    & TVN & 11.600 ± 0.000 & 5.361 ± 0.005 & 5.960 ± 0.019 \\
    \midrule
    \multirow{4}{*}{Transfer scVI} 
    & Center Scaling & 10.600 ± 0.000 & 5.285 ± 0.006 & 5.470 ± 0.005 \\
    & Centering & 10.700 ± 0.000 & 5.285 ± 0.006 & 5.477 ± 0.010 \\
    & Raw Embeds & 4.800 ± 0.000 & 5.282 ± 0.006 & 5.450 ± 0.006 \\
    & TVN & 11.800 ± 0.000 & 5.282 ± 0.006 & 5.399 ± 0.011 \\
    \midrule
    \multirow{4}{*}{UCE} 
    & Center Scaling & 9.800 ± 0.000 & 5.288 ± 0.007 & 5.499 ± 0.008 \\
    & Centering & 9.500 ± 0.000 & 5.288 ± 0.007 & 5.495 ± 0.008 \\
    & Raw Embeds & 4.600 ± 0.000 & 5.288 ± 0.007 & 5.485 ± 0.006 \\
    & TVN & 10.500 ± 0.000 & 5.297 ± 0.009 & 5.397 ± 0.011 \\
    \midrule
    \multirow{4}{*}{cellPLM} 
    & Center Scaling & 5.800 ± 0.000 & 5.291 ± 0.007 & 5.537 ± 0.004 \\
    & Centering & 6.000 ± 0.000 & 5.292 ± 0.007 & 5.568 ± 0.006 \\
    & Raw Embeds & 5.400 ± 0.000 & 5.291 ± 0.007 & 5.544 ± 0.004 \\
    & TVN & 11.100 ± 0.000 & 5.301 ± 0.005 & 5.483 ± 0.010 \\
    \midrule
    \multirow{4}{*}{scGPT} 
    & Center Scaling & 9.700 ± 0.000 & 5.299 ± 0.007 & 5.577 ± 0.009 \\
    & Centering & 10.200 ± 0.000 & 5.299 ± 0.007 & 5.580 ± 0.012 \\
    & Raw Embeds & 4.100 ± 0.000 & 5.295 ± 0.007 & 5.530 ± 0.008 \\
    & TVN & 11.500 ± 0.000 & 5.314 ± 0.006 & 5.478 ± 0.009 \\
    \midrule
    \multirow{4}{*}{scGPT finetuned} 
    & Center Scaling & 6.200 ± 0.000 & 5.282 ± 0.006 & 5.374 ± 0.008 \\
    & Centering & 6.500 ± 0.000 & 5.280 ± 0.006 & 5.375 ± 0.007 \\
    & Raw Embeds & 6.000 ± 0.000 & 5.280 ± 0.006 & 5.381 ± 0.007 \\
    & TVN & 9.200 ± 0.000 & 5.280 ± 0.006 & 5.372 ± 0.009 \\
    \midrule
    \multirow{4}{*}{scVI} 
    & Center Scaling & 9.400 ± 0.000 & 5.367 ± 0.011 & 6.133 ± 0.018 \\
    & Centering & 9.500 ± 0.000 & 5.366 ± 0.009 & 6.126 ± 0.005 \\
    & Raw Embeds & 10.200 ± 0.000 & 5.356 ± 0.009 & 6.028 ± 0.016 \\
    & TVN & 9.600 ± 0.000 & 5.309 ± 0.009 & 5.622 ± 0.015 \\
    \bottomrule
    \end{tabular}
\end{table}


\begin{table}[ht]
\centering
\caption{\cite{Replogle2022-lx} dataset full results}
\label{tab:performance_metrics}
\resizebox{\textwidth}{!}{%
\begin{tabular}{llccccc}
\toprule
\textbf{Model Type} & \textbf{Processing Method} & \textbf{CORUM} & \textbf{HuMAP} & \textbf{Reactome} & \textbf{SIGNOR} & \textbf{StringDB} \\
\midrule
\multirow{4}{*}{Geneformer} 
& Center Scaling & 0.099 ± 0.000 & 0.103 ± 0.000 & 0.117 ± 0.000 & 0.108 ± 0.000 & 0.108 ± 0.000 \\
& Centering & 0.096 ± 0.000 & 0.097 ± 0.000 & 0.109 ± 0.000 & 0.106 ± 0.000 & 0.103 ± 0.000 \\
& Raw Embeds & 0.149 ± 0.000 & 0.135 ± 0.000 & 0.106 ± 0.000 & 0.108 ± 0.000 & 0.148 ± 0.000 \\
& TVN & 0.099 ± 0.000 & 0.099 ± 0.000 & 0.101 ± 0.000 & 0.100 ± 0.000 & 0.102 ± 0.000 \\
\midrule
\multirow{4}{*}{PCA} 
& Center Scaling & 0.450 ± 0.000 & 0.359 ± 0.000 & 0.213 ± 0.000 & 0.135 ± 0.000 & 0.475 ± 0.000 \\
& Centering & 0.257 ± 0.000 & 0.249 ± 0.000 & 0.136 ± 0.000 & 0.150 ± 0.000 & 0.307 ± 0.000 \\
& Raw Embeds & 0.202 ± 0.000 & 0.200 ± 0.000 & 0.125 ± 0.000 & 0.137 ± 0.000 & 0.247 ± 0.000 \\
& TVN & 0.425 ± 0.000 & 0.356 ± 0.000 & 0.202 ± 0.000 & 0.124 ± 0.000 & 0.461 ± 0.000 \\
\midrule
\multirow{4}{*}{Transfer scVI} 
& Center Scaling & 0.443 ± 0.000 & 0.344 ± 0.000 & 0.172 ± 0.000 & 0.122 ± 0.000 & 0.423 ± 0.000 \\
& Centering & 0.443 ± 0.000 & 0.342 ± 0.000 & 0.171 ± 0.000 & 0.123 ± 0.000 & 0.420 ± 0.000 \\
& Raw Embeds & 0.138 ± 0.000 & 0.163 ± 0.000 & 0.087 ± 0.000 & 0.082 ± 0.000 & 0.144 ± 0.000 \\
& TVN & 0.420 ± 0.000 & 0.346 ± 0.000 & 0.174 ± 0.000 & 0.115 ± 0.000 & 0.428 ± 0.000 \\
\midrule
\multirow{4}{*}{UCE} 
& Center Scaling & 0.366 ± 0.000 & 0.253 ± 0.000 & 0.170 ± 0.000 & 0.128 ± 0.000 & 0.329 ± 0.000 \\
& Centering & 0.344 ± 0.000 & 0.231 ± 0.000 & 0.162 ± 0.000 & 0.126 ± 0.000 & 0.306 ± 0.000 \\
& Raw Embeds & 0.132 ± 0.000 & 0.102 ± 0.000 & 0.097 ± 0.000 & 0.096 ± 0.000 & 0.111 ± 0.000 \\
& TVN & 0.285 ± 0.000 & 0.249 ± 0.000 & 0.141 ± 0.000 & 0.115 ± 0.000 & 0.309 ± 0.000 \\
\midrule
\multirow{4}{*}{cellPLM} 
& Center Scaling & 0.250 ± 0.000 & 0.263 ± 0.000 & 0.107 ± 0.000 & 0.102 ± 0.000 & 0.284 ± 0.000 \\
& Centering & 0.271 ± 0.000 & 0.272 ± 0.000 & 0.113 ± 0.000 & 0.106 ± 0.000 & 0.299 ± 0.000 \\
& Raw Embeds & 0.128 ± 0.000 & 0.230 ± 0.000 & 0.070 ± 0.000 & 0.078 ± 0.000 & 0.156 ± 0.000 \\
& TVN & 0.284 ± 0.000 & 0.259 ± 0.000 & 0.144 ± 0.000 & 0.121 ± 0.000 & 0.305 ± 0.000 \\
\midrule
\multirow{4}{*}{scGPT} 
& Center Scaling & 0.389 ± 0.000 & 0.285 ± 0.000 & 0.161 ± 0.000 & 0.123 ± 0.000 & 0.373 ± 0.000 \\
& Centering & 0.371 ± 0.000 & 0.263 ± 0.000 & 0.155 ± 0.000 & 0.123 ± 0.000 & 0.353 ± 0.000 \\
& Raw Embeds & 0.121 ± 0.000 & 0.107 ± 0.000 & 0.093 ± 0.000 & 0.103 ± 0.000 & 0.108 ± 0.000 \\
& TVN & 0.383 ± 0.000 & 0.323 ± 0.000 & 0.166 ± 0.000 & 0.116 ± 0.000 & 0.402 ± 0.000 \\
\midrule
\multirow{4}{*}{scGPT finetuned} 
& Center Scaling & 0.126 ± 0.000 & 0.112 ± 0.000 & 0.110 ± 0.000 & 0.111 ± 0.000 & 0.131 ± 0.000 \\
& Centering & 0.119 ± 0.000 & 0.111 ± 0.000 & 0.106 ± 0.000 & 0.110 ± 0.000 & 0.126 ± 0.000 \\
& Raw Embeds & 0.143 ± 0.000 & 0.135 ± 0.000 & 0.098 ± 0.000 & 0.094 ± 0.000 & 0.139 ± 0.000 \\
& TVN & 0.166 ± 0.000 & 0.149 ± 0.000 & 0.116 ± 0.000 & 0.110 ± 0.000 & 0.170 ± 0.000 \\
\midrule
\multirow{4}{*}{scVI} 
& Center Scaling & 0.503 ± 0.000 & 0.377 ± 0.000 & 0.225 ± 0.000 & 0.146 ± 0.000 & 0.494 ± 0.000 \\
& Centering & 0.501 ± 0.000 & 0.377 ± 0.000 & 0.224 ± 0.000 & 0.147 ± 0.000 & 0.492 ± 0.000 \\
& Raw Embeds & 0.224 ± 0.000 & 0.211 ± 0.000 & 0.103 ± 0.000 & 0.084 ± 0.000 & 0.243 ± 0.000 \\
& TVN & 0.502 ± 0.000 & 0.388 ± 0.000 & 0.219 ± 0.000 & 0.137 ± 0.000 & 0.497 ± 0.000 \\
\midrule
\multirow{1}{*}{Rand Labels} 
& Raw Embeds & 0.107 ± 0.018 & 0.102 ± 0.012 & 0.090 ± 0.008 & 0.097 ± 0.013 & 0.102 ± 0.012 \\
\bottomrule
\end{tabular}
} 
\end{table}

\begin{table}[ht]
    \centering
    \caption{\cite{Replogle2022-lx} dataset full results}
    \label{tab:spearman_structural}
    \begin{tabular}{llcc}
    \toprule
    \textbf{Model Type} & \textbf{Processing Method} & \textbf{Spearman Corr} & \textbf{Structural Integrity} \\
    \midrule
    \multirow{4}{*}{Geneformer} 
    & Center Scaling & 0.152 ± 0.0016 & 0.523 ± 0.0107 \\
    & Centering      & 0.150 ± 0.0046 & 0.526 ± 0.0119 \\
    & Raw Embeds     & 0.149 ± 0.0020 & 0.528 ± 0.0105 \\
    & TVN            & -0.0003 ± 0.0010 & 0.520 ± 0.0114 \\
    \midrule
    \multirow{4}{*}{PCA} 
    & Center Scaling & 0.187 ± 0.0019 & 0.548 ± 0.0099 \\
    & Centering      & 0.207 ± 0.0021 & 0.548 ± 0.0110 \\
    & Raw Embeds     & 0.215 ± 0.0021 & 0.551 ± 0.0098 \\
    & TVN            & 0.196 ± 0.0024 & 0.546 ± 0.0110 \\
    \midrule
    \multirow{4}{*}{Transfer scVI} 
    & Center Scaling & 0.202 ± 0.0025 & 0.547 ± 0.0099 \\
    & Centering      & 0.205 ± 0.0022 & 0.548 ± 0.0099 \\
    & Raw Embeds     & 0.218 ± 0.0023 & 0.549 ± 0.0098 \\
    & TVN            & 0.203 ± 0.0026 & 0.545 ± 0.0111 \\
    \midrule
    \multirow{4}{*}{UCE} 
    & Center Scaling & 0.151 ± 0.0018 & 0.534 ± 0.0116 \\
    & Centering      & 0.172 ± 0.0030 & 0.537 ± 0.0113 \\
    & Raw Embeds     & 0.170 ± 0.0026 & 0.538 ± 0.0101 \\
    & TVN            & 0.142 ± 0.0022 & 0.533 ± 0.0117 \\
    \midrule
    \multirow{4}{*}{cellPLM} 
    & Center Scaling & 0.160 ± 0.0029 & 0.535 ± 0.0108 \\
    & Centering      & 0.164 ± 0.0032 & 0.537 ± 0.0111 \\
    & Raw Embeds     & 0.169 ± 0.0033 & 0.538 ± 0.0101 \\
    & TVN            & 0.167 ± 0.0028 & 0.537 ± 0.0113 \\
    \midrule
    \multirow{4}{*}{scGPT} 
    & Center Scaling & 0.168 ± 0.0026 & 0.539 ± 0.0101 \\
    & Centering      & 0.180 ± 0.0030 & 0.540 ± 0.0101 \\
    & Raw Embeds     & 0.180 ± 0.0029 & 0.540 ± 0.0101 \\
    & TVN            & 0.159 ± 0.0031 & 0.536 ± 0.0114 \\
    \midrule
    \multirow{4}{*}{scGPT finetuned} 
    & Center Scaling & 0.098 ± 0.0026 & 0.529 ± 0.0104 \\
    & Centering      & 0.131 ± 0.0034 & 0.532 ± 0.0102 \\
    & Raw Embeds     & 0.128 ± 0.0034 & 0.532 ± 0.0102 \\
    & TVN            & 0.084 ± 0.0021 & 0.525 ± 0.0117 \\
    \midrule
    \multirow{4}{*}{scVI} 
    & Center Scaling & 0.247 ± 0.0016 & 0.549 ± 0.0098 \\
    & Centering      & 0.250 ± 0.0015 & 0.549 ± 0.0099 \\
    & Raw Embeds     & 0.265 ± 0.0015 & 0.551 ± 0.0098 \\
    & TVN            & 0.254 ± 0.0016 & 0.547 ± 0.0110 \\
    \midrule
    \multirow{1}{*}{Rand Labels} 
    & Raw Embeds     & 0.001 ± 0.0005 & 0.526 ± 0.0111 \\
    \bottomrule
    \end{tabular}
\end{table}

\begin{table}[ht]
    \centering
    \caption{L1000 dataset full results}
    \label{tab:performance_metrics}
    \resizebox{\textwidth}{!}{%
    \begin{tabular}{llccc}
    \toprule
    \textbf{Model Type} & \textbf{Processing Method} & \textbf{iLISI} & \textbf{Top1 Linear Acc} & \textbf{Top5 Linear Acc} \\
    \midrule
    \multirow{4}{*}{Geneformer} 
    & Center Scaling & 26.151 ± 0.021 & 0.071 ± 0.004 & 0.279 ± 0.014 \\
    & Centering & 26.510 ± 0.015 & 0.138 ± 0.031 & 0.536 ± 0.073 \\
    & Raw Embeds & 25.010 ± 0.022 & 0.833 ± 0.010 & 2.197 ± 0.015 \\
    & TVN & 6.094 ± 0.050 & 0.091 ± 0.004 & 0.432 ± 0.016 \\
    \midrule
    \multirow{4}{*}{PCA} 
    & Center Scaling & 15.276 ± 0.294 & 2.691 ± 0.029 & 5.225 ± 0.044 \\
    & Centering & 13.760 ± 0.038 & 2.628 ± 0.019 & 5.328 ± 0.030 \\
    & Raw Embeds & 1.519 ± 0.000 & 2.067 ± 0.021 & 4.920 ± 0.048 \\
    & TVN & 2.295 ± 0.024 & 1.745 ± 0.035 & 3.914 ± 0.025 \\
    \midrule
    \multirow{4}{*}{Transfer scVI} 
    & Center Scaling & 17.934 ± 0.069 & 0.662 ± 0.009 & 1.693 ± 0.019 \\
    & Centering & 14.630 ± 0.165 & 0.683 ± 0.017 & 1.819 ± 0.035 \\
    & Raw Embeds & 2.122 ± 0.001 & 1.019 ± 0.011 & 2.580 ± 0.017 \\
    & TVN & 2.870 ± 0.007 & 0.506 ± 0.007 & 1.521 ± 0.034 \\
    \midrule
    \multirow{4}{*}{UCE} 
    & Center Scaling & 26.303 ± 0.021 & 0.042 ± 0.003 & 0.187 ± 0.011 \\
    & Centering & 26.744 ± 0.019 & 0.073 ± 0.005 & 0.353 ± 0.015 \\
    & Raw Embeds & 26.820 ± 0.011 & 0.833 ± 0.010 & 2.183 ± 0.015 \\
    & TVN & 25.916 ± 0.092 & 0.075 ± 0.006 & 0.293 ± 0.009 \\
    \midrule
    \multirow{4}{*}{cellPLM} 
    & Center Scaling & 2.802 ± 0.014 & 0.469 ± 0.034 & 0.942 ± 0.034 \\
    & Centering & 2.512 ± 0.005 & 0.717 ± 0.008 & 1.002 ± 0.014 \\
    & Raw Embeds & 3.861 ± 0.001 & 0.830 ± 0.010 & 2.187 ± 0.013 \\
    & TVN & 1.795 ± 0.008 & 0.482 ± 0.018 & 1.883 ± 0.043 \\
    \midrule
    \multirow{4}{*}{scGPT} 
    & Center Scaling & 25.327 ± 0.029 & 0.159 ± 0.014 & 0.528 ± 0.015 \\
    & Centering & 25.297 ± 0.026 & 0.188 ± 0.036 & 0.636 ± 0.074 \\
    & Raw Embeds & 11.346 ± 0.011 & 0.834 ± 0.010 & 2.190 ± 0.015 \\
    & TVN & 5.206 ± 0.037 & 0.121 ± 0.010 & 0.484 ± 0.009 \\
    \midrule
    \multirow{4}{*}{scGPT finetuned} 
    & Center Scaling & 25.771 ± 0.039 & 0.126 ± 0.003 & 0.393 ± 0.011 \\
    & Centering & 26.187 ± 0.036 & 0.158 ± 0.081 & 0.568 ± 0.157 \\
    & Raw Embeds & 22.416 ± 0.030 & 0.833 ± 0.010 & 2.196 ± 0.015 \\
    & TVN & 6.155 ± 0.047 & 0.115 ± 0.008 & 0.501 ± 0.012 \\
    \midrule
    \multirow{4}{*}{scVI} 
    & Center Scaling & 15.171 ± 0.173 & 2.121 ± 0.034 & 4.370 ± 0.022 \\
    & Centering & 15.650 ± 0.108 & 2.105 ± 0.038 & 4.336 ± 0.027 \\
    & Raw Embeds & 2.249 ± 0.002 & 1.837 ± 0.030 & 4.205 ± 0.046 \\
    & TVN & 2.503 ± 0.026 & 1.565 ± 0.041 & 3.608 ± 0.014 \\
    \bottomrule
    \end{tabular}
    } 
\end{table}

\begin{table}[ht]
    \centering
    \caption{L1000 dataset full results}
    \label{tab:performance_metrics}
    \resizebox{\textwidth}{!}{%
    \begin{tabular}{llccc}
    \toprule
    \textbf{Model Type} & \textbf{Processing Method} & \textbf{Pert Consistency} & \textbf{Top1 KNN Acc} & \textbf{Top5 KNN Acc} \\
    \midrule
    \multirow{4}{*}{Geneformer} 
    & Center Scaling & 5.900 ± 0.000 & 25.605 ± 0.111 & 26.180 ± 0.088 \\
    & Centering & 4.800 ± 0.000 & 25.605 ± 0.111 & 26.183 ± 0.085 \\
    & Raw Embeds & 4.600 ± 0.000 & 25.605 ± 0.111 & 26.181 ± 0.103 \\
    & TVN & 5.700 ± 0.000 & 25.606 ± 0.111 & 25.978 ± 0.115 \\
    \midrule
    \multirow{4}{*}{PCA} 
    & Center Scaling & 7.300 ± 0.000 & 25.605 ± 0.111 & 26.733 ± 0.123 \\
    & Centering & 4.900 ± 0.000 & 25.605 ± 0.111 & 26.752 ± 0.094 \\
    & Raw Embeds & 6.100 ± 0.000 & 25.605 ± 0.111 & 27.102 ± 0.096 \\
    & TVN & 6.800 ± 0.000 & 25.605 ± 0.111 & 26.510 ± 0.102 \\
    \midrule
    \multirow{4}{*}{Transfer scVI} 
    & Center Scaling & 6.100 ± 0.000 & 25.605 ± 0.111 & 26.350 ± 0.084 \\
    & Centering & 6.000 ± 0.000 & 25.605 ± 0.111 & 26.380 ± 0.103 \\
    & Raw Embeds & 4.100 ± 0.000 & 25.605 ± 0.111 & 26.855 ± 0.107 \\
    & TVN & 6.400 ± 0.000 & 25.605 ± 0.111 & 26.296 ± 0.100 \\
    \midrule
    \multirow{4}{*}{UCE} 
    & Center Scaling & 5.800 ± 0.000 & 25.605 ± 0.111 & 26.303 ± 0.021 \\
    & Centering & 5.600 ± 0.000 & 25.605 ± 0.111 & 26.744 ± 0.019 \\
    & Raw Embeds & 4.600 ± 0.000 & 25.605 ± 0.111 & 26.820 ± 0.011 \\
    & TVN & 5.500 ± 0.000 & 25.613 ± 0.112 & 25.916 ± 0.092 \\
    \midrule
    \multirow{4}{*}{cellPLM} 
    & Center Scaling & 6.800 ± 0.000 & 25.600 ± 0.113 & 26.244 ± 0.101 \\
    & Centering & 5.800 ± 0.000 & 25.605 ± 0.111 & 26.233 ± 0.112 \\
    & Raw Embeds & 5.800 ± 0.000 & 25.609 ± 0.111 & 26.887 ± 0.107 \\
    & TVN & 2.900 ± 0.000 & 25.778 ± 0.122 & 27.032 ± 0.115 \\
    \midrule
    \multirow{4}{*}{scGPT} 
    & Center Scaling & 5.100 ± 0.000 & 25.605 ± 0.111 & 26.209 ± 0.109 \\
    & Centering & 4.700 ± 0.000 & 25.605 ± 0.111 & 26.150 ± 0.096 \\
    & Raw Embeds & 3.200 ± 0.000 & 25.605 ± 0.111 & 26.345 ± 0.113 \\
    & TVN & 5.600 ± 0.000 & 25.605 ± 0.111 & 25.986 ± 0.124 \\
    \midrule
    \multirow{4}{*}{scGPT finetuned} 
    & Center Scaling & 5.400 ± 0.000 & 25.605 ± 0.111 & 26.155 ± 0.107 \\
    & Centering & 5.800 ± 0.000 & 25.605 ± 0.111 & 26.143 ± 0.092 \\
    & Raw Embeds & 4.500 ± 0.000 & 25.605 ± 0.111 & 26.184 ± 0.120 \\
    & TVN & 5.600 ± 0.000 & 25.607 ± 0.113 & 26.003 ± 0.130 \\
    \midrule
    \multirow{4}{*}{scVI} 
    & Center Scaling & 7.800 ± 0.000 & 25.605 ± 0.111 & 26.651 ± 0.104 \\
    & Centering & 8.200 ± 0.000 & 25.605 ± 0.111 & 26.581 ± 0.092 \\
    & Raw Embeds & 3.400 ± 0.000 & 25.605 ± 0.111 & 26.780 ± 0.103 \\
    & TVN & 7.300 ± 0.000 & 25.605 ± 0.111 & 26.489 ± 0.133 \\
    \midrule
    \multirow{1}{*}{Rand Labels} 
    & Raw Embeds & 4.740 ± 1.137 & 25.605 ± 0.111 & 38.344 ± 0.055 \\
    \bottomrule
    \end{tabular}
    } 
\end{table}

\begin{table}[ht]
    \centering
    \caption{L1000 dataset full results}
    \label{tab:performance_metrics}
    \resizebox{\textwidth}{!}{%
    \begin{tabular}{llccccc}
    \toprule
    \textbf{Model Type} & \textbf{Processing Method} & \textbf{CORUM} & \textbf{HuMAP} & \textbf{Reactome} & \textbf{SIGNOR} & \textbf{StringDB} \\
    \midrule
    \multirow{4}{*}{Geneformer} 
    & Center Scaling & 0.108 ± 0.000 & 0.109 ± 0.000 & 0.113 ± 0.000 & 0.095 ± 0.000 & 0.115 ± 0.000 \\
    & Centering & 0.108 ± 0.000 & 0.106 ± 0.000 & 0.110 ± 0.000 & 0.091 ± 0.000 & 0.117 ± 0.000 \\
    & Raw Embeds & 0.141 ± 0.000 & 0.124 ± 0.000 & 0.128 ± 0.000 & 0.145 ± 0.000 & 0.150 ± 0.000 \\
    & TVN & 0.095 ± 0.000 & 0.114 ± 0.000 & 0.111 ± 0.000 & 0.096 ± 0.000 & 0.122 ± 0.000 \\
    \midrule
    \multirow{4}{*}{PCA} 
    & Center Scaling & 0.174 ± 0.000 & 0.153 ± 0.000 & 0.115 ± 0.000 & 0.120 ± 0.000 & 0.143 ± 0.000 \\
    & Centering & 0.154 ± 0.000 & 0.132 ± 0.000 & 0.109 ± 0.000 & 0.139 ± 0.000 & 0.134 ± 0.000 \\
    & Raw Embeds & 0.112 ± 0.000 & 0.124 ± 0.000 & 0.133 ± 0.000 & 0.104 ± 0.000 & 0.157 ± 0.000 \\
    & TVN & 0.123 ± 0.000 & 0.134 ± 0.000 & 0.144 ± 0.000 & 0.099 ± 0.000 & 0.161 ± 0.000 \\
    \midrule
    \multirow{4}{*}{Transfer scVI} 
    & Center Scaling & 0.201 ± 0.000 & 0.168 ± 0.000 & 0.130 ± 0.000 & 0.164 ± 0.000 & 0.163 ± 0.000 \\
    & Centering & 0.203 ± 0.000 & 0.173 ± 0.000 & 0.133 ± 0.000 & 0.161 ± 0.000 & 0.165 ± 0.000 \\
    & Raw Embeds & 0.096 ± 0.000 & 0.114 ± 0.000 & 0.130 ± 0.000 & 0.082 ± 0.000 & 0.147 ± 0.000 \\
    & TVN & 0.112 ± 0.000 & 0.126 ± 0.000 & 0.141 ± 0.000 & 0.101 ± 0.000 & 0.161 ± 0.000 \\
    \midrule
    \multirow{4}{*}{UCE} 
    & Center Scaling & 0.098 ± 0.000 & 0.110 ± 0.000 & 0.104 ± 0.000 & 0.107 ± 0.000 & 0.103 ± 0.000 \\
    & Centering & 0.095 ± 0.000 & 0.106 ± 0.000 & 0.101 ± 0.000 & 0.105 ± 0.000 & 0.103 ± 0.000 \\
    & Raw Embeds & 0.163 ± 0.000 & 0.121 ± 0.000 & 0.141 ± 0.000 & 0.204 ± 0.000 & 0.146 ± 0.000 \\
    & TVN & 0.092 ± 0.000 & 0.099 ± 0.000 & 0.099 ± 0.000 & 0.098 ± 0.000 & 0.097 ± 0.000 \\
    \midrule
    \multirow{4}{*}{cellPLM} 
    & Center Scaling & 0.111 ± 0.000 & 0.111 ± 0.000 & 0.107 ± 0.000 & 0.105 ± 0.000 & 0.106 ± 0.000 \\
    & Centering & 0.101 ± 0.000 & 0.091 ± 0.000 & 0.102 ± 0.000 & 0.101 ± 0.000 & 0.107 ± 0.000 \\
    & Raw Embeds & 0.105 ± 0.000 & 0.104 ± 0.000 & 0.104 ± 0.000 & 0.114 ± 0.000 & 0.106 ± 0.000 \\
    & TVN & 0.074 ± 0.000 & 0.116 ± 0.000 & 0.123 ± 0.000 & 0.080 ± 0.000 & 0.147 ± 0.000 \\
    \midrule
    \multirow{4}{*}{scGPT} 
    & Center Scaling & 0.160 ± 0.000 & 0.146 ± 0.000 & 0.111 ± 0.000 & 0.128 ± 0.000 & 0.136 ± 0.000 \\
    & Centering & 0.156 ± 0.000 & 0.147 ± 0.000 & 0.114 ± 0.000 & 0.126 ± 0.000 & 0.137 ± 0.000 \\
    & Raw Embeds & 0.090 ± 0.000 & 0.099 ± 0.000 & 0.113 ± 0.000 & 0.085 ± 0.000 & 0.130 ± 0.000 \\
    & TVN & 0.102 ± 0.000 & 0.105 ± 0.000 & 0.124 ± 0.000 & 0.095 ± 0.000 & 0.135 ± 0.000 \\
    \midrule
    \multirow{4}{*}{scGPT finetuned} 
    & Center Scaling & 0.103 ± 0.000 & 0.107 ± 0.000 & 0.089 ± 0.000 & 0.094 ± 0.000 & 0.108 ± 0.000 \\
    & Centering & 0.097 ± 0.000 & 0.107 ± 0.000 & 0.088 ± 0.000 & 0.098 ± 0.000 & 0.106 ± 0.000 \\
    & Raw Embeds & 0.140 ± 0.000 & 0.100 ± 0.000 & 0.087 ± 0.000 & 0.082 ± 0.000 & 0.147 ± 0.000 \\
    & TVN & 0.117 ± 0.000 & 0.100 ± 0.000 & 0.112 ± 0.000 & 0.097 ± 0.000 & 0.127 ± 0.000 \\
    \midrule
    \multirow{4}{*}{scVI} 
    & Center Scaling & 0.182 ± 0.000 & 0.158 ± 0.000 & 0.122 ± 0.000 & 0.134 ± 0.000 & 0.151 ± 0.000 \\
    & Centering & 0.190 ± 0.000 & 0.170 ± 0.000 & 0.127 ± 0.000 & 0.129 ± 0.000 & 0.161 ± 0.000 \\
    & Raw Embeds & 0.132 ± 0.000 & 0.123 ± 0.000 & 0.149 ± 0.000 & 0.119 ± 0.000 & 0.159 ± 0.000 \\
    & TVN & 0.109 ± 0.000 & 0.130 ± 0.000 & 0.137 ± 0.000 & 0.084 ± 0.000 & 0.159 ± 0.000 \\
    \midrule
    \multirow{1}{*}{Rand Labels} 
    & Raw Embeds & 0.111 ± 0.008 & 0.113 ± 0.008 & 0.109 ± 0.005 & 0.120 ± 0.011 & 0.109 ± 0.002 \\
    \bottomrule
    \end{tabular}
    } 
\end{table}

\begin{table}[ht]
    \centering
    \caption{L1000 dataset full results}
    \label{tab:spearman_structural}
    \begin{tabular}{llcc}
    \toprule
    \textbf{Model Type} & \textbf{Processing Method} & \textbf{Spearman Corr} & \textbf{Structural Integrity} \\
    \midrule
    \multirow{4}{*}{Geneformer} 
    & Center Scaling & 0.402 ± 0.007 & 0.938 ± 0.003 \\
    & Centering & 0.382 ± 0.005 & 0.943 ± 0.002 \\
    & Raw Embeds & 0.734 ± 0.003 & 0.955 ± 0.002 \\
    & TVN & 0.748 ± 0.003 & 0.954 ± 0.003 \\
    \midrule
    \multirow{4}{*}{PCA} 
    & Center Scaling & 0.717 ± 0.005 & 0.953 ± 0.002 \\
    & Centering & 0.730 ± 0.013 & 0.954 ± 0.003 \\
    & Raw Embeds & 0.882 ± 0.004 & 0.974 ± 0.001 \\
    & TVN & 0.882 ± 0.003 & 0.971 ± 0.002 \\
    \midrule
    \multirow{4}{*}{Transfer scVI} 
    & Center Scaling & 0.669 ± 0.009 & 0.948 ± 0.002 \\
    & Centering & 0.764 ± 0.005 & 0.955 ± 0.002 \\
    & Raw Embeds & 0.903 ± 0.001 & 0.970 ± 0.002 \\
    & TVN & 0.821 ± 0.004 & 0.962 ± 0.002 \\
    \midrule
    \multirow{4}{*}{UCE} 
    & Center Scaling & 0.202 ± 0.015 & 0.934 ± 0.003 \\
    & Centering & 0.217 ± 0.006 & 0.942 ± 0.002 \\
    & Raw Embeds & 0.236 ± 0.009 & 0.942 ± 0.002 \\
    & TVN & 0.250 ± 0.024 & 0.928 ± 0.004 \\
    \midrule
    \multirow{4}{*}{cellPLM} 
    & Center Scaling & 0.462 ± 0.017 & 0.944 ± 0.003 \\
    & Centering & 0.440 ± 0.004 & 0.943 ± 0.002 \\
    & Raw Embeds & 0.812 ± 0.002 & 0.960 ± 0.002 \\
    & TVN & 0.592 ± 0.224 & 0.938 ± 0.025 \\
    \midrule
    \multirow{4}{*}{scGPT} 
    & Center Scaling & 0.715 ± 0.004 & 0.953 ± 0.002 \\
    & Centering & 0.690 ± 0.006 & 0.954 ± 0.002 \\
    & Raw Embeds & 0.833 ± 0.002 & 0.963 ± 0.002 \\
    & TVN & 0.766 ± 0.004 & 0.957 ± 0.003 \\
    \midrule
    \multirow{4}{*}{scGPT finetuned} 
    & Center Scaling & 0.700 ± 0.007 & 0.951 ± 0.003 \\
    & Centering & 0.461 ± 0.006 & 0.946 ± 0.002 \\
    & Raw Embeds & 0.636 ± 0.008 & 0.951 ± 0.002 \\
    & TVN & 0.758 ± 0.005 & 0.956 ± 0.003 \\
    \midrule
    \multirow{4}{*}{scVI} 
    & Center Scaling & 0.744 ± 0.008 & 0.955 ± 0.002 \\
    & Centering & 0.778 ± 0.006 & 0.958 ± 0.002 \\
    & Raw Embeds & 0.928 ± 0.002 & 0.977 ± 0.001 \\
    & TVN & 0.881 ± 0.004 & 0.971 ± 0.002 \\
    \midrule
    \multirow{1}{*}{Rand Labels} 
    & Raw Embeds & 0.001 ± 0.001 & 0.939 ± 0.002 \\
    \bottomrule
    \end{tabular}
\end{table}



\end{document}